\documentclass[sigconf]{acmart}

\usepackage{algorithm2e}
\usepackage{hyperref}
\usepackage{multirow}
\usepackage{graphicx}
\usepackage{subfigure}
\usepackage{balance}

\theoremstyle{plain}
\newtheorem{theorem}{Theorem}

\AtBeginDocument{%
}

\copyrightyear{2025}
\acmYear{2025}
\setcopyright{rightsretained}
\acmConference[KDD '25]{Proceedings of the 31st ACM SIGKDD Conference on Knowledge Discovery and Data Mining V.1}{August 3--7, 2025}{Toronto, ON, Canada} \acmBooktitle{Proceedings of the 31st ACM SIGKDD Conference on Knowledge Discovery and Data Mining V.1 (KDD '25), August 3--7, 2025, Toronto, ON, Canada} \acmDOI{10.1145/3690624.3709173} \acmISBN{979-8-4007-1245-6/25/08}

\makeatletter
\gdef\@copyrightpermission{
  \begin{minipage}{0.2\columnwidth}
   \href{https://creativecommons.org/licenses/by-nc-sa/4.0/}{\includegraphics[width=0.90\textwidth]{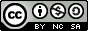}}
  \end{minipage}\hfill
  \begin{minipage}{0.8\columnwidth}
   \href{https://creativecommons.org/licenses/by-nc-sa/4.0/}{This work is licensed under a Creative Commons Attribution-NonCommercial-ShareAlike International 4.0 License.}
  \end{minipage}
  \vspace{5pt}
}
\makeatother

\begin{document}

\title{Harnessing Scale and Physics: A Multi-Graph Neural Operator Framework for PDEs on Arbitrary Geometries}

\author{Zhihao Li}
\orcid{0000-0003-4752-6811}
\affiliation{%
  \institution{The Hong Kong University of Science and Technology (Guangzhou)}
  \city{Guangzhou}
  \country{China}
}
\email{zli416@connect.hkust-gz.edu.cn}

\author{Haoze Song}
\affiliation{%
  \institution{The Hong Kong University of Science and Technology (Guangzhou)}
  \city{Guangzhou}
  \country{China}
}
\email{hsong492@connect.hkust-gz.edu.cn}

\author{Di Xiao}
\affiliation{%
  \institution{Beihang University}
  \city{Beijing}
  \country{China}
}
\email{shawd@buaa.edu.cn}

\author{Zhilu Lai}
\affiliation{
    \institution{The Hong Kong University of Science and Technology (Guangzhou)}
    \city{Guangzhou}
    \country{China}\\
    \institution{The Hong Kong University of Science and Technology}
    \city{Hong Kong SAR}
    \country{China}
}
\email{zhilulai@ust.hk}

\author{Wei Wang}
\authornote{Corresponding author.}
\affiliation{
    \institution{The Hong Kong University of Science and Technology (Guangzhou)}
    \city{Guangzhou}
    \country{China}\\
    \institution{The Hong Kong University of Science and Technology}
    \city{Hong Kong SAR}
    \country{China}
}
\email{weiwcs@ust.hk}

\renewcommand{\shortauthors}{Zhihao Li, Haoze Song, Di Xiao, Zhilu Lai, \& Wei Wang}

\begin{abstract}
Partial Differential Equations (PDEs) underpin many scientific phenomena, yet traditional computational approaches often struggle with complex, nonlinear systems and irregular geometries. This paper introduces the \textbf{AMG} method, a \textbf{M}ulti-\textbf{G}raph neural operator approach designed for efficiently solving PDEs on \textbf{A}rbitrary geometries. AMG leverages advanced graph-based techniques and dynamic attention mechanisms within a novel GraphFormer architecture, enabling precise management of diverse spatial domains and complex data interdependencies. By constructing multi-scale graphs to handle variable feature frequencies and a physics graph to encapsulate inherent physical properties, AMG significantly outperforms previous methods, which are typically limited to uniform grids. We present a comprehensive evaluation of AMG across six benchmarks, demonstrating its consistent superiority over existing state-of-the-art models. Our findings highlight the transformative potential of tailored graph neural operators in surmounting the challenges faced by conventional PDE solvers. Our code and datasets are available on \url{https://github.com/lizhihao2022/AMG}.
\end{abstract}




\maketitle

\newcommand\kddavailabilityurl{https://doi.org/10.5281/zenodo.14542736}

\ifdefempty{\kddavailabilityurl}{}{
\begingroup\small\noindent\raggedright\textbf{KDD Availability Link:}\\
The source code of this paper has been made publicly available at \url{\kddavailabilityurl}.
\endgroup
}

\begin{figure*}[ht]
    \centering
    \includegraphics[width=\textwidth]{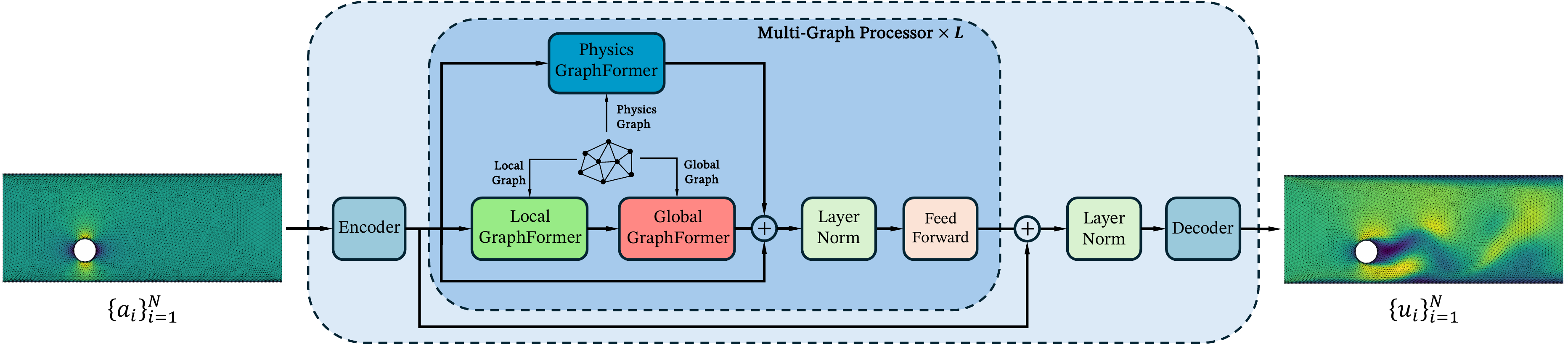}
    \caption{Overview of the model architecture.}
    \Description{Overview of the model architecture.}
    \label{fig:arc}
\end{figure*}

\section{Introduction}
Partial Differential Equations (PDEs) underlie critical phenomena across diverse fields, from fluid dynamics to quantum mechanics, showcasing their ability to model complex variable relationships. Traditional approaches, while foundational, often struggle with the complexities and non-linearities of these systems, particularly in irregular geometries. Operator Learning has emerged as a transformative approach, leveraging deep learning architectures like Deep Operator Networks (DeepONet) and Fourier Neural Operators (FNO) to directly map input conditions to PDE solutions \cite{21:deeponet,21:fno,23:no}. This method stands out for its versatility, requiring no retraining for different conditions, thus enabling efficient model adaptation across diverse settings.

However, applying these deep learning techniques to real-world problems poses significant challenges, especially with irregular geometries where traditional methods like FNO \cite{21:fno}, using Fast Fourier Transform, and U-Net \cite{15:UNet}, employing convolutions, are inherently limited to uniform grids. Innovations such as Graph Neural Operators (GNO) \cite{20:GNO}, and geometric adaptations like Geo-FNO \cite{24:GeoFNO}, attempt to address these limitations by projecting irregular domains into more manageable latent meshes. Moreover, GINO combines these approaches to enhance modeling across various scales \cite{23:GINO}. However, the inherent limitations of Fourier bases, particularly under the periodic boundary assumption, lead to significant performance degradation in complex geometries \cite{77:Numerical}. Similarly, graph kernels often fail to capture global information effectively \cite{23:GNOT, 23:LSM, 24:Transolver}.

The challenge extends to the domain's frequency spectrum; high-frequency areas require more intensive learning efforts compared to the predominantly flat, low-frequency regions. This disparity in learning demand across different frequencies, akin to techniques used in computer vision for tasks such as super-resolution, underscores the potential of tailoring neural network structures to leverage these differences for enhanced performance \cite{23:RethinkingImg,18:LearningDC}.

In PDE learning, multi-scale methods have shown promise; however, they are generally limited to uniform grids and often require dimensions to be powers of two, particularly in methods utilizing wavelets \cite{20:MGNO,21:MWT,24:M2NO}. Furthermore, transformer-based models, which distribute sample points across equal-sized attention grids, struggle with the highly variable frequency behaviors seen in phenomena like wave propagation \cite{21:Choose, 23:GNOT, 23:LSM, 24:Transolver}. Existing graph-based methods, which often employ a $k$-Nearest-Neighbor approach to construct graphs, treat all nodes equally, neglecting the diverse learning needs inherent in PDE dynamics. This one-size-fits-all approach to node degrees can significantly impede the efficacy of graph-based models.

To address the challenges associated with solving PDEs on arbitrary geometries and capturing multi-scale features, we introduce \textbf{AMG}—a method employing a \textbf{M}ulti-scale \textbf{G}raph neural operator tailored for \textbf{A}rbitrary geometries. AMG leverages three distinct types of graphs, constructed from the given coordinates: two are designed for multi-scale processing is dedicated to capturing the underlying physical properties. At the heart of AMG is the GraphFormer, a novel architecture equipped with dynamic graph attention mechanisms. This design allows the model to compute the hidden representations of each node by dynamically attending to its neighbors, effectively handling the complex interdependencies within the data. Furthermore, we theoretically establish that graph attention can be seen as a learnable integral, enhancing our method's ability to model continuous spaces. Extensive experiments were conducted on six benchmarks, including four well-established ones and two custom-designed to test our method's efficacy across diverse geometries and dynamic mesh configurations. AMG consistently outperforms existing solutions, achieving a remarkable relative gain on six benchmarks.

The contributions of this paper are summarized as follows:
\begin{itemize}
    \item We use local sampling (Section \ref{subsec:local_graph}) and global sampling (Section \ref{subsec:global_graph}) to construct multi-scale graphs for capturing different frequencies of features and a physical graph (Section \ref{subsec:phy_graph}) to encode inherent physical properties effectively.
    \item We introduce a GraphFormer (Section \ref{subsec:graphformer}) architecture with dynamic graph attention mechanisms, providing a scalable and flexible encoder-processor-decoder framework (Section \ref{subsec:overview}) for learning operators tailored to arbitrary geometries.
    \item AMG demonstrates superior performance, achieving consistent state-of-the-art results with significant relative gains across a variety of benchmarks.
\end{itemize}

\section{Preliminaries}

\subsection{Problem Formulation}
We consider Partial Differential Equations (PDEs) defined over a domain \(D \subset \mathbb{R}^d\). Define \(\mathcal{A} = \mathcal{A}(D; \mathbb{R}^{d_a})\) and \(\mathcal{U} = \mathcal{U}(D; \mathbb{R}^{d_u})\) as two Sobolev spaces \(\mathcal{H}^{s,p}\), with parameters \(s > 0\) and \(p \geq 1\). Our aim is to learn an operator \(\mathcal{G} : \mathcal{A} \to \mathcal{U}\), mapping from the input function space \(\mathcal{A}\) to the solution function space \(\mathcal{U}\). Specifically, we select \(s > 0\) and \(p = 2\) to take advantage of the Hilbert space structure, which facilitates the definition of projections.

The operator \(\mathcal{G}\) is characterized as an integral operator with a kernel $\kappa$, where $\kappa: D \times D \to L^2$, and is formalized by the integral equation:
\begin{equation}
    \mathcal{G}a(\mathbf{x}) = \int_{D} \kappa(\mathbf{x}, \mathbf{y}) a(\mathbf{y}) \, d\mathbf{y},
\end{equation}
enabling \(\mathcal{G}\) to operate within the Hilbert space framework and leveraging the properties of \(L^2\) spaces for enhanced analytical and computational efficiency.

\subsection{Graph Neural Operators}
A directed graph \(\mathcal{G}=(\mathcal{V}, \mathcal{E})\) includes nodes \(\mathcal{V}=\{1,...,n\}\) and edges \(\mathcal{E} \subseteq \mathcal{V} \times \mathcal{V}\), with \((j,i) \in \mathcal{E}\) representing an edge from node \(j\) to node \(i\). Each node \(i \in \mathcal{V}\) initially possesses a representation \(\mathbf{h}_i^{(0)} \in \mathbb{R}^{d_{h}}\). An undirected graph is depicted using bidirectional edges between nodes.

A Graph Neural Network (GNN) layer updates the representation of each node by aggregating information from its neighbors. The input to a GNN layer consists of a set of node representations \(\{\mathbf{h}_i \in \mathbb{R}^{d_{h}} \mid i \in \mathcal{V}\}\) and the set of edges \(\mathcal{E}\). The output is a new set of node representations \(\{\mathbf{h}'_i \in \mathbb{R}^{d_{h}} \mid i \in \mathcal{V}\}\), where each node's updated state is calculated as:
\begin{equation}
    \mathbf{h}'_i = f_{\theta}(\mathbf{h}_i, \mathrm{\text{AGGREGATE}}(\{\mathbf{h}_j \mid j \in \mathcal{N}_i\}))
\end{equation}
The function \(f\) and the aggregation method \(\mathrm{\text{AGGREGATE}}\) largely define the distinctions among various GNN architectures.

If we accurately construct the graph on the spatial domain \(D\) of the PDE, the kernel integration can be interpreted as an aggregation of messages \cite{20:GNO}. In the edge-conditioned aggregation mechanism \cite{17:NeuralMP, 17:DynamicEC} utilized at the \(k^{\text{th}}\) layer, the output feature \(\mathbf{h}_{i}^{k}\) for node \(i\) is calculated based on its previous feature \(\mathbf{h}^{k-1}\) as follows:
\begin{equation}
    u(\mathbf{x}_{i}) = \mathbf{h}_{i}^{k} = \frac{\sum_{j \in \mathcal{N}(i)} \exp{(\kappa(\mathbf{x}_j, \mathbf{x}_i)) \mathbf{h}_{j}^{k-1}}}{\sum_{j \in \mathcal{N}(i)} \exp{(\kappa(\mathbf{x}_j, \mathbf{x}_i))}},
\end{equation}
where \(\kappa : \mathbb{R}^{d} \times \mathbb{R}^{d} \to \mathbb{R}\) is a parameterized kernel function that evaluates the correlation between node pairs \((j, i)\), influencing the strength of the message passed from node \(j\) to node \(i\).

\section{Methodology}
This section outlines the AMG method's architecture for solving PDEs on arbitrary geometries. We start with an architecture overview in Section \ref{subsec:overview}, followed by the construction of multi-graphs in Section \ref{subsec:construc}. The processing module is detailed in Section \ref{subsec:processor}, and the GraphFormer Block is discussed in Section \ref{subsec:graphformer}. 

\subsection{Overview of Model Architecture} \label{subsec:overview}
Our model architecture is designed to effectively handle the complexities involved in processing partial differential equations (PDEs) on irregular geometries using graph neural networks. The architecture is illustrated in Figure \ref{fig:arc}.

\textbf{Input and Encoder:} The process begins with the input dataset \(\{a_i\}_{i=1}^N\), which consists of the initial conditions or parameters for the PDEs. These inputs are first processed by an Encoder, which transforms the raw data into a preliminary feature representation suitable for further processing within the neural network.

\textbf{Graph Construction:} Two distinct types of graphs, the Physics Graph and the Multi-Scale Graph, are constructed from the encoded features. The Physics Graph encapsulates the underlying physical laws governing the phenomena being modeled, while the Multi-Scale Graph captures interactions at various scales, crucial for accurately modeling complex systems.

\textbf{Multi-Graph Processing Layers:} The core of the architecture comprises multiple processing layers, each consisting of a GraphFormer, Message Passing, and Layer Normalization components followed by a Feed Forward network. The GraphFormer component applies transformations specific to graph data, facilitating the propagation and update of node features. Message Passing enables the exchange of information between nodes, enhancing the model's ability to learn from the topology of the graph. Layer Normalization is employed to stabilize the learning process, and the Feed Forward networks provide additional transformation capabilities to the node features.

\textbf{Decoder and Output:} Following the processing layers, a Decoder reverts the graph-based features back into the spatial domain, generating the output \(\{u_i\}_{i=1}^N\) which represents the solution to the PDE at the discretized points. This output effectively demonstrates the model's capability to predict complex phenomena governed by the PDEs.

This architecture leverages the strengths of graph neural networks to process data over irregular domains, ensuring robustness and accuracy in capturing the dynamics of the system modeled.

\subsection{Graph Construction} \label{subsec:construc}
\subsubsection{High-Frequency Indicator}
To efficiently and effectively pinpoint regions within feature maps that contain detail-rich information, we introduce a high-frequency indicator. This indicator is designed to rapidly identify high-frequency areas that are crucial for accurate PDE solutions in complex geometries. Given a feature map of point set \( F \in \mathbb{R}^{N \times C} \) ,where $C$ is the number of feature channels and a specific down-sampling ratio \( s \), the high frequency indicator per node, \( H_{F} \in \mathbb{R}^{N} \), is computed as follows:
\begin{equation} \label{eq:HF}
    H_{F} = \sum_{c=1}^C \left| F^{(c)} - (F^{(c)})_{\downarrow s \uparrow s} \right|,
\end{equation}
where \( (F^{(c)})_{\downarrow s \uparrow s} \) denotes the channel \( c \) of the feature map \( F \) after it has been bilinearly down-sampled and subsequently up-sampled by a factor of \( s \). The choice of \( s \) effectively balances detail preservation with minimal information loss, optimizing the process for quick and effective high-frequency detection.

\subsubsection{Partitioning of Point Sets} \label{subsec:fps}
To efficiently manage the challenge of generating overlapping partitions within a point set, we define each partition as a neighborhood ball within Euclidean space, characterized by centroid location and scale. We utilize a farthest point sampling (FPS) algorithm \cite{17:PointNetDH} to select centroids in a manner that ensures even coverage across the entire point set. This method starts with an arbitrary initial point and iteratively selects subsequent points that are the farthest in metric distance from all previously selected points. This approach not only guarantees superior coverage compared to random sampling but also takes into account the spatial distribution of the points, contrasting with methods like convolutional neural networks (CNNs) that process data agnostically of spatial distributions. Consequently, our graph construction strategy establishes receptive fields in a data-dependent manner, significantly enhancing the model's capability to capture and process complex spatial relationships. A detailed description of the FPS algorithm is provided in Appendix \ref{appendix:fps}.

\subsubsection{Local Sampling} \label{subsec:local_graph}
Local sampling is essential for constructing graphs that precisely capture the immediate neighborhood dynamics of a node, particularly valuable in systems characterized by complex local interactions, such as in solving PDEs with intricate local behaviors. This method selects each node \( v \) and its neighbors \( \mathcal{N}(v) \) based on a high-frequency indicator (Eq.\eqref{eq:HF}), identifying nodes that reside within areas of high detail and information density.

In practice, nodes are chosen for their rich detail using the high-frequency indicator to determine the composition of each local graph. For every node \( v \), its neighbors \( \mathcal{N}(v) \) are selected to ensure that only nodes within areas of substantial detail and information content are included. This targeted selection strategy guarantees that the connections within the graph are meaningful and representative of significant local interactions. By focusing on nodes with high information content, the model can adapt more effectively to variations in data density and local structural complexities. This approach is invaluable in scenarios demanding high precision in local detail, enabling the model to accurately capture complex dynamics.

To maintain local coherence, we use the Euclidean distance between two nodes as the metric for establishing connections:
\begin{equation}
    d_{\mathrm{local}}(i,j) = \| \mathbf{p}_{i} - \mathbf{p}_{j} \|_{2},
\end{equation}
where \(\mathbf{p}_i\) and \(\mathbf{p}_j\) are the position vectors of nodes \(i\) and \(j\), respectively. This metric also allows for connecting nodes that belong to different domains.

An illustrative example of this local sampling process and the resultant graph structure is shown in Figure \ref{fig:graph}(b). This visualization underscores how local sampling concentrates on areas rich in detail, thereby substantially enhancing the model's capacity to learn nuanced behaviors and interactions, which are crucial for solving complex PDEs. A detailed description of the local sampling algorithm is provided in Appendix \ref{appendix:local}.

\subsubsection{Global Sampling} \label{subsec:global_graph}
Conversely, global sampling aims to capture the broader structure of the dataset by establishing connections across wider spatial extents. This sampling strategy selects nodes that span the entire domain in a pattern designed to maximize coverage and minimize redundancy, typically implemented using a dilated sampling technique as introduced in Section \ref{subsec:fps}. By increasing the intervals between selected nodes, this method provides a comprehensive overview of global interactions essential for understanding large-scale trends and phenomena.

Connectivity among nodes is established through a metric that measures the similarity in node features, employing cosine similarity:
\begin{equation}
    d_{\mathrm{global}}(i, j) = \frac{\mathbf{h}_{i} \cdot \mathbf{h}_{j}}{\| \mathbf{h}_{i} \| \| \mathbf{h}_{j} \|},
\end{equation}
where \( \mathbf{h}_i \) and \( \mathbf{h}_j \) are the feature vectors of nodes \( i \) and \( j \), respectively. This metric ensures that nodes with similar features are linked, regardless of their physical placement, thus enhancing the graph's ability to model phenomena accurately. 

An illustrative example of this global sampling process is shown in Figure \ref{fig:graph}(b). The combination of both local and global sampling strategies in graph construction allows the model to effectively balance detail-oriented and holistic learning objectives.

\subsubsection{Physics Graph} \label{subsec:phy_graph}
In our model, node attributes are embedded into higher dimensions, allowing us to interpret them as high-level physical attributes, which are often the solutions sought in complex physical systems. These high-level attributes derive from more fundamental, lower-level physical properties.

To effectively represent these relationships, we construct a phy-sics graph where each node corresponds to a fundamental physical attribute. In this graph, nodes from the standard operational graph are linked to all nodes in the physics graph, emphasizing the foundational contributions of these attributes to higher-level phenomena. The edges within the physics graph symbolize the interactions between each of these lower-level physical attributes.

In scenarios where explicit physical information about the connections between attributes is available, the physics graph is constructed based on this empirical data. However, in cases where such specifics are lacking, we opt for a fully-connected graph configuration. This approach is justified by the typically small number of lower-level nodes and the frequent interactions among them, ensuring comprehensive coverage of potential influences and interactions within the system.

\begin{figure}[ht]
    \centering
    \subfigure[]{\includegraphics[width=0.32\linewidth]{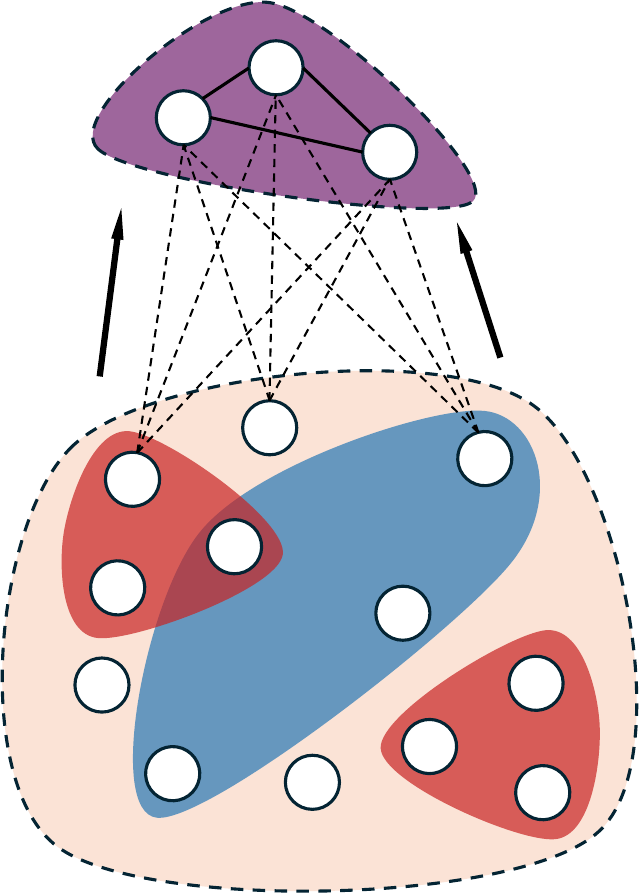}}\quad\quad
    \subfigure[]{\includegraphics[width=0.20\linewidth]{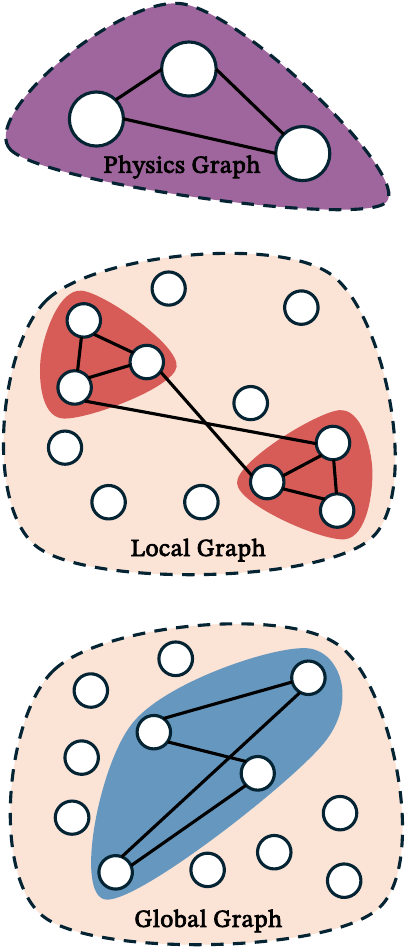}}\quad\quad
    \subfigure[]{\includegraphics[width=0.32\linewidth]{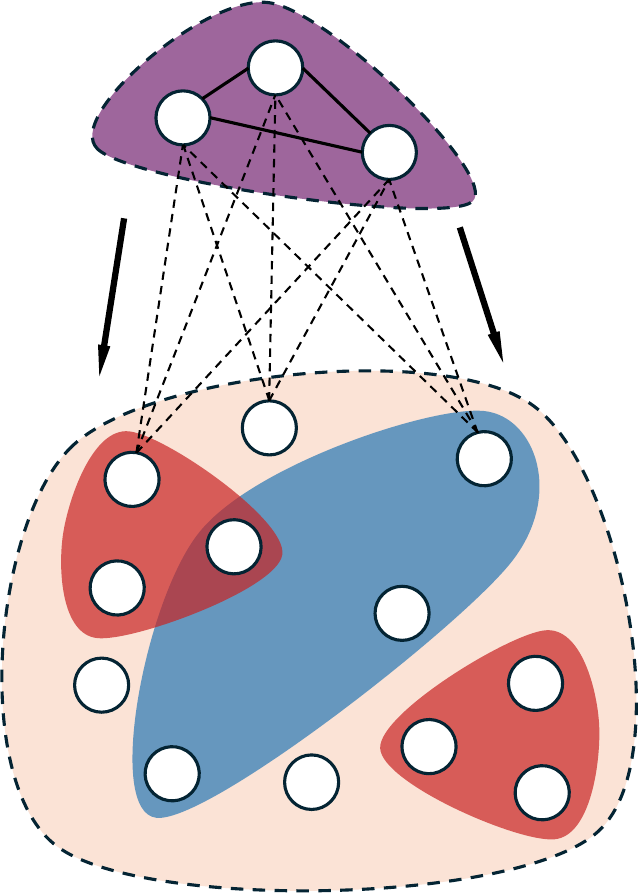}}
    \caption{(a) Message passing from multi-scale graph to physics graph. (b) Three types of graphs as the input of GraphFormer. (c) Message passing from physics graph to multi-scale graph.}
    \Description{}
    \label{fig:graph}
\end{figure}

An illustration of the construction of the physics graph is shown in Figure \ref{fig:graph}(a), and the detailed propagation mechanisms between the physics graph and the multi-scale graph are defined in Section \ref{subsec:phy_prop}.

\subsection{Multi-Graph Processor} \label{subsec:processor}

Multi-Scale Graph Blocks integrate both local and global-scale information crucial for effective operator learning. These blocks compute local and global graphs per block based on the inputs, allowing for dynamic graph updates throughout the model's processing.

The design of the Multi-Scale Graph Blocks follows the general MetaFormer structure \cite{22:MetaFormer}, employing GraphFormers to perform graph aggregation acting as token mixers. Depending on the type of graph received, a GraphFormer can process either local or global information, enhancing its flexibility and capability to handle complex data structures.

\subsubsection{Cross-Scale Graph Aggregation}
After constructing two \\graphs at different scales, as shown in Figure \ref{fig:graph}(b), nodes are processed through the GraphFormer Block, defining local or global edges accordingly:
\begin{align}
 \{\mathbf{h}_{i}^{local}\}_{i=1}^{N} &= \text{GraphFormer}(\{\mathbf{h}_{i}^{x}\}_{i=1}^{N}, \mathcal{E}_{local}),\\
    \{\mathbf{h}_{i}^{global}\}_{i=1}^{N} &= \text{GraphFormer}(\{\mathbf{h}_{i}^{local}\}_{i=1}^{N}, \mathcal{E}_{global}),
\end{align}
where $\mathbf{h}_{i}^{x}$ represents the input features for node $i$, and $\mathcal{E}_{local}$ and $\mathcal{E}_{global}$ are the sets of edges in the local and global graphs, respectively.

\subsubsection{Physics Graph Propagation} \label{subsec:phy_prop}
As depicted in Figure \ref{fig:graph}(a), virtual physics nodes are initialized by aggregating information from the original graph nodes. Specifically, the value of virtual physics nodes $\{\mathbf{h}_{j}^{v}\}_{j=1}^{M}$ is calculated as:
\begin{equation}
    \mathbf{h}_{j}^{v}=\sum_{i=1}^{N}{\mathbf{e}_{ij}^{v}\mathbf{h}_{i}^{x}},
\end{equation}
where $\mathbf{h}_{i}^{x}$ are the nodes from the original graph, and $\mathbf{e}_{ij}^{v}$ are the edge weights defined by:
\begin{equation}
    \mathbf{e}_{ij}^{v}=\frac{\mathbf{W}_{v}\mathbf{h}_{i}^{x}}{\sum_{i'\in\mathcal{N}_{j}^{v}}{\mathbf{h}_{i'}^{x}}},
\end{equation}
with $\mathbf{W}_{v} \in \mathbb{R}^{d_{h} \times d_{h}}$ being a trainable linear layer. Subsequently, the transformed physical nodes are processed through the GraphFormer:
\begin{equation}
    \{\mathbf{h'}_{j}^{v}\}_{j=1}^{M} = \text{GraphFormer}(\{\mathbf{h}_{j}^{v}\}_{j=1}^{M}, \mathcal{E}_{phy}),
\end{equation}
Finally, these physical nodes are reintegrated into the graph nodes through a message-passing scheme, effectively blending the computed physical node representations back into the global node features:
\begin{equation}
    \mathbf{h'}_{i}^{x}=\mathbf{h}_{i}^{global} + \sum_{j=1}^{M}{\mathbf{e}_{ij}^{v}\mathbf{h'}_{j}^{v}},
\end{equation}
where each token $\mathbf{h'}_{j}^{v}$ is broadcasted to all graph nodes during the calculation. This method ensures that both local and global information is incorporated into each node, facilitating a comprehensive learning of the PDE operators across different scales and physical interpretations.

\subsection{GraphFormer Block} \label{subsec:graphformer}

\begin{figure}[ht]
    \centering
    \includegraphics[width=\linewidth]{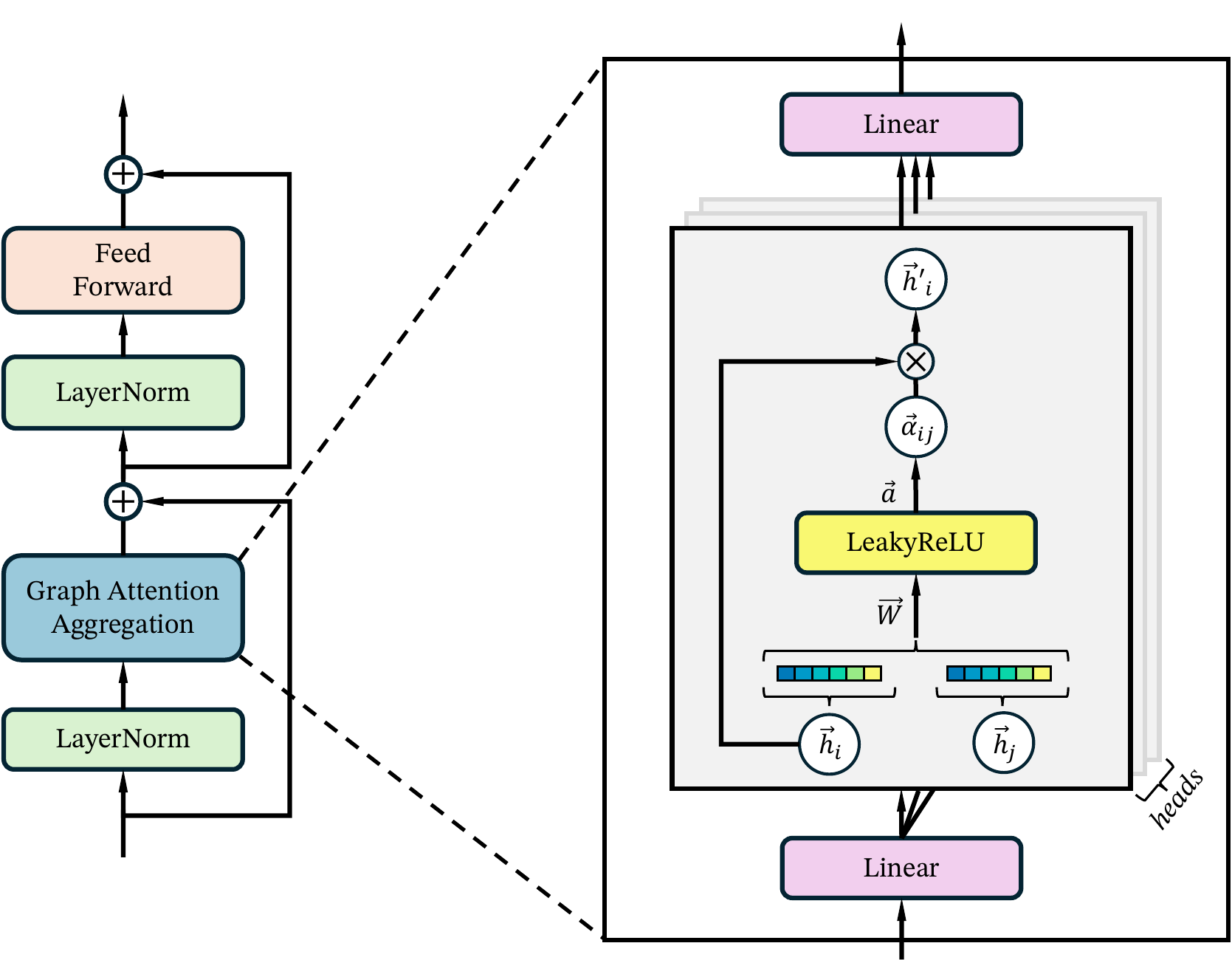}
    \caption{\textbf{Left:} The Architecture of GraphFormer Block. \textbf{Right:} An illustration of multi-head attention by node $i$ on its neighborhood $j$.}
    \Description{Grapher Block}
    \label{fig:graphfomer}
    \vspace{-5pt}
\end{figure}

Inspired by the general MetaFormer architecture \cite{22:MetaFormer}, we introduce the GraphFormer block, which replaces traditional attention mechanisms with graph attention, thereby tailoring the approach to handle graph-based data more effectively. As depicted in Figure \ref{fig:graphfomer}, this block employs graph aggregation techniques as token mixers, facilitating efficient information processing across the network.


\subsubsection{Graph Attention Mechanism}
A scoring function \( f : \mathbb{R}^d_{h} \times \mathbb{R}^d_{h} \rightarrow \mathbb{R} \) evaluates the importance of features from neighbor \( j \) to node \( i \) in every edge \((j, i)\):
\begin{equation}
    f(\mathbf{h}_i, \mathbf{h}_j) = \text{LeakyReLU} \left( \mathbf{a}^\top \cdot \left[ \mathbf{W}\mathbf{h}_i \| \mathbf{W}\mathbf{h}_j \right] \right),
\end{equation}
where \( \mathbf{a} \in \mathbb{R}^{2d_{h}} \) and \( \mathbf{W} \in \mathbb{R}^{d_{h} \times d_{h}} \) are trainable parameters, and \(\| \) denotes vector concatenation. The attention scores are then normalized using a softmax function across all neighbors \( j \in \mathcal{N}_i \):
\begin{equation} \label{eq:alpha}
    \alpha_{ij} = \text{softmax}_j(f(\mathbf{h}_i, \mathbf{h}_j)) = \frac{\exp(f(\mathbf{h}_i, \mathbf{h}_j))}{\sum_{j' \in \mathcal{N}_i} \exp(f(\mathbf{h}_i, \mathbf{h}_{j'}))}.
\end{equation}
The Graph Attention (GA) then computes the new representation for node \( i \) as a weighted average of transformed features of its neighbors, applying a nonlinearity \( \sigma \):
\begin{equation} \label{eq:ga}
    \mathbf{h}_i' = \sigma \left( \sum_{j \in \mathcal{N}_i} \alpha_{ij} \cdot \mathbf{W}\mathbf{h}_j \right).
\end{equation}
Following \cite{22:GATv2}, we apply the \( \mathbf{a} \) layer after the LeakyReLU non-linearity and \( \mathbf{W} \) after concatenation, essentially using an MLP to compute the score for each query-key pair. 

The overall complexity is $\mathcal{O}(|\mathcal{V}|d_{h}^{2}+|\mathcal{E}|d_{h})$, where  $|\mathcal{V}|$ represents the number of nodes and $|\mathcal{E}|$ the number of edges in the graphs used, either multi-scale or physics. Since $|\mathcal{V}| \ll N$  and $|\mathcal{E}| = k|\mathcal{V}| \leq |\mathcal{V}|^2$ , the computational complexity remains linear with respect to the number of mesh points $N$. This ensures that the proposed method scales efficiently even as the mesh complexity increases.

\subsubsection{Properties of GraphFormer}
Prior methods have approached PDE learning as an iterative update process, demonstrating that canonical attention mechanisms can approximate integral operators over the input domain \( \Omega \) \cite{21:Choose, 23:no}. To deepen our theoretical understanding of the GraphFormer, we propose that it similarly acts as a learnable integral on \( \Omega \):

\begin{theorem}[GraphFormer as a Learnable Integral on \( \Omega \)] \label{th:integral}
    Given an input function \( a: \Omega \to \mathbb{R}^{d} \) and a mesh point \( \mathbf{x} \in \Omega \), GraphFormer seeks to approximate the integral operator \( \mathcal{G} \), defined by:
    \begin{equation}
        \mathcal{G}a(\mathbf{x}) = \int_{\Omega} \kappa(\mathbf{x}, \boldsymbol{\xi}) a(\boldsymbol{\xi}) \, d\boldsymbol{\xi},
    \end{equation}
    where \( \kappa(\cdot, \cdot) \) denotes a kernel function over \( \Omega \times \Omega \).
\end{theorem}

A complete proof is available in Appendix \ref{appendix:proof}. This theoretical foundation substantiates the capability of GraphFormer to efficiently learn complex mappings essential for solving PDEs across irregular domains, underscoring its utility in advanced computational mathematics.


\section{Experiments}
\subsection{General Setting}
In practice, to learn a neural operator, we use a dataset $\mathcal{D} = \{(a, u)\}$, where $u = \mathcal{G}(a)$. Due to challenges in directly representing functions, we discretize the input and solution functions over the domain $D$ using an irregular mesh, as per a specified mesh generation algorithm \cite{98:mesh}. Thus, we consider a set of function pairs $(a_i, u_i)_{i=1}^{N}$, where $a_i = a(\mathbf{x}_i)$ and $u_i = u(\mathbf{x}_i)$ at discretized points $\mathbf{x}_i$ within the domain $D\subset\mathbb{R}^{d}$.

Our goal is to approximate $\mathcal{G}$ by optimizing the network parameters $\theta$, through the following optimization problem:
\begin{equation}
    \min_{\theta \in \Theta} \mathcal{L}(\theta) := \min_{\theta \in \Theta} \frac{1}{N} \sum_{i=1}^{N} \left[ \lVert \tilde{\mathcal{G}}_{\theta}(a_i) - u_i \rVert^2 \right].
\end{equation}


\begin{table}[ht]
    \caption{Summary of Experiment Benchmarks.}
    \label{table:bm}
    \centering
    \renewcommand{\arraystretch}{1.4}  
    \setlength{\tabcolsep}{4.7pt}       
    \resizebox{0.98\linewidth}{!}{
    \begin{tabular}{c|c|c|c|c|c}
        \toprule
        \textbf{Mesh Type} & \textbf{Benchmarks} & \textbf{Geometry} & \# \textbf{Dim} & \# \textbf{Mesh} & \# \textbf{TimeStep} \\
        \midrule
        \multirow{5}{*}{Fixed} 
            & Navier-Stokes    & Structured    & 2 & 4096          & 10    \\
            & Shape-Net Car    & Unstructured  & 3 & 32186         & /     \\
            & Poisson           & Unstructured  & 2 & 3242          & /     \\
            & Airfoil           & Unstructured  & 2 & 5233          & 10    \\
            & Weather Forecast  & Unstructured  & 2 & 11088         & 30    \\
        \midrule 
        \multirow{2}{*}{Dynamic} 
            & Cylinder Flow     & Unstructured  & 2 & 7209 $\sim$ 7581 & 100  \\
            & Deforming Plate   & Unstructured  & 3 & 672 $\sim$ 2189   & 50   \\
        \bottomrule
    \end{tabular}}
\end{table}

\begin{table*}[ht]
    \caption{Main results on benchmarks. A smaller value indicates better performance. For clarity, the best result is in \textbf{bold} and the second best is \underline{underlined}.}
    \label{table:main}
    \centering
    \resizebox{\linewidth}{!}{
    \begin{tabular}{c|cccccccccccc}
        \toprule
        \multirow{2}{*}{Model} & \multirow{2}{*}{NS} & \multicolumn{2}{c}{Shape-Net Car} & \multirow{2}{*}{Poisson} & \multicolumn{4}{c}{Airfoil} & \multirow{2}{*}{DP} & \multicolumn{3}{c}{Cylinder Flow} \\
        \cmidrule(lr){3-4} \cmidrule(lr){6-9} \cmidrule(lr){11-13}
         & & pressure & velocity & & density & pressure & velocity x & velocity y & & pressure & velocity x & velocity y \\
        \midrule
        MLP & 0.1607 & 0.2422 & 0.2428 & 0.4804 & 0.0963 & 0.0947 & 0.0438 & 0.0559 & 0.0856 & 0.0781 & 0.0447 & 0.1753 \\
        U-Net & 0.0970 & 0.1380 & 0.1867 & 0.2699 & 0.0579 & 0.0575 & 0.0536 & 0.0523 & 0.0873 & 0.0558 & 0.0224 & 0.0562 \\
        \midrule
        MGN & 0.1614 & 0.2432 & 0.2477 & 0.4428 & 0.0506 & 0.0491 & 0.0208 & 0.0258 & 0.0323 & 0.0774 & 0.0456 & 0.1715 \\
        GNO & 0.1610 & 0.2453 & 0.2596 & 0.5167 & 0.0786 & 0.0769 & 0.0286 & 0.0406 & 0.0414 & 0.0855 & 0.0509 & 0.1835 \\
        \midrule
        ONO & 0.0824 & 0.1115 & 0.1389 & 0.0199 & 0.0097 & 0.0096 & 0.0075 & 0.0088 & 0.0358 & 0.0201 & 0.0157 & 0.0263 \\
        GNOT & 0.5805 & 0.1109 & \underline{0.1206} & 0.4403 & 0.0054 & 0.0049 & 0.0044 & 0.0040 & 0.2362 & 0.0392 & 0.0272 & 0.0545 \\
        LSM & 0.0957 & 0.1169 & 0.1487 & 0.2612 & 0.0305 & 0.0299 & 0.0525 & 0.0341 & 0.0802 & 0.0093 & 0.0065 & 0.0126 \\
        Transolver & \underline{0.0593} & \underline{0.0993} & 0.1208 & \underline{0.0162} & \underline{0.0036} & \underline{0.0032} & \underline{0.0028} & \underline{0.0035} & \underline{0.0264} & \underline{0.0092} & \underline{0.0059} & \underline{0.0117} \\
        \midrule
        \textbf{AMG (Ours)} & \textbf{0.0476} & \textbf{0.0878} & \textbf{0.0919} & \textbf{0.0152} & \textbf{0.0021} & \textbf{0.0020} & \textbf{0.0014} & \textbf{0.0018} & \textbf{0.0257} & \textbf{0.0050} & \textbf{0.0038} & \textbf{0.0078} \\
        Promotion & 19.70\% & 11.64\% & 23.81\% & 6.44\% & 39.97\% & 38.01\% & 51.44\% & 47.45\% & 2.35\% & 45.33\% & 35.44\% & 32.94\% \\
        \bottomrule
    \end{tabular}}
\end{table*}

\subsubsection{Benchmarks}
We conducted comprehensive analyses across a variety of benchmark scenarios to demonstrate the superiority of our method, as detailed in \cite{21:fno,21:learningMS,24:GeoFNO}. These benchmarks were carefully selected to encompass a wide range of PDE problems, from fluid dynamics to structural deformations, using both static and dynamic conditions. Specifically, we developed datasets for the Poisson equation and Cylinder Flow, testing our model on both fixed and dynamic unstructured meshes, as well as standard structured meshes. We also expand our experimental scope to include global weather forecasting, utilizing data from the European Centre for Medium-Range Weather Forecasts (ECMWF). This new experiment involves climate data from the first quarter of 2018, focusing specifically on the atmospheric conditions at a pressure level of 50hPa. This breadth of testing highlights our model's adaptability and precision across various computational environments. For ease of reference, these benchmarks are summarized in Table \ref{table:bm}, with visual examples provided and detailed configurations discussed in Appendix \ref{appendix:bm}.

\subsubsection{Baselines}
We conducted a comprehensive evaluation against a variety of established methods to validate our approach. Traditional models like MLP \cite{86:mlp} and U-Net \cite{15:UNet} were included for their foundational roles in computational tasks. We also assessed against specialized graph-based methods such as MeshGraphNets (MGN) \cite{21:learningMS} and Graph Neural Operator (GNO) \cite{20:GNO}, which are designed for complex spatial data. Additionally, our model was compared with advanced transformer-based methods, including LSM \cite{23:LSM}, GNOT \cite{23:GNOT}, ONO \cite{23:ONO} and Transolver \cite{24:Transolver}, to benchmark against the latest approaches in handling high-dimensional data complexities.

\subsubsection{Implementation}
To ensure fairness in performance comparison, all models were uniformly trained across 500 epochs using the \(L_2\) loss function \cite{21:fno}. We utilized the Adam \cite{14:adam} and AdamW \cite{19:AdamW} optimizers, initiating with a learning rate of \(10^{-3}\) and implementing a decay factor of \(\gamma=0.5\) to halve the rate every 100 epochs. For handling dynamic mesh sizes, which transformer-based models inherently struggle with, we adopted a padding strategy similar to that used in GNOT \cite{23:GNOT} to accommodate these models. All experiments were conducted using a single Nvidia A800 80GB GPU, and unless specified, default hyperparameters were employed for the baseline models. This setup facilitated a controlled environment to accurately assess the comparative effectiveness of each method. 

\subsection{Main Results}
AMG's robustness and accuracy across diverse PDE benchmarks are highlighted in the performance comparison (Table \ref{table:main}), demonstrating significant enhancements in both structured and unstructured mesh environments.

\subsubsection{Structured Mesh Performance}
In the Navier-Stokes benchmark with structured 2D geometry, AMG achieved a notable reduction in errors by 19.70\%. This illustrates AMG's effectiveness in structured environments where regular geometries predominate, requiring high precision to accurately model dynamic interactions. 

\subsubsection{Fixed Unstructured Mesh Analysis}
In the Poisson benchmark, AMG improved prediction by 6.44\%, demonstrating its effectiveness in modeling static spatial variations in simpler systems. 
In the Shape-Net Car benchmark involving complex 3D automobile geometries, AMG achieved improvements of 11.64\% in pressure accuracy of surface and 23.81\% in velocity accuracy of surrounding air, showcasing its capability to handle intricate 3D shapes and multiple attributes in a large unstructured mesh environment.
For the Airfoil benchmark, AMG excelled by enhancing velocity predictions by 51.44\% for velocity X and 47.45\% for velocity Y, along with increases of 39.97\% in density and 38.01\% in pressure accuracy. This demonstrates its proficiency in dynamic simulations, capturing detailed airflow dynamics and interactions. 
Overall, these results affirm AMG's robust performance across diverse fixed unstructured mesh scenarios, illustrating its versatility and high precision in handling complex physical phenomena across different dimensions and conditions.

\subsubsection{Dynamic Unstructured Mesh Challenges}
AMG's prowess was distinctly displayed in the Cylinder Flow and Deforming Plate benchmarks, both demanding high adaptability in dynamic environments. 
In the 2D Cylinder Flow scenario, AMG significantly enhanced performance metrics, improving pressure accuracy by 45.33\%, velocity X by 35.44\%, and velocity Y by 32.94\% over 100 timesteps. These enhancements confirm AMG's exceptional ability to handle temporal changes and adaptive mesh akin to real-world fluid dynamics.
The 3D Deforming Plate benchmark posed additional challenges with both spatial and temporal dynamics. Here, AMG achieved a 2.35\% improvement in pressure accuracy, demonstrating its competency in managing the intricacies of 3D dynamic systems that necessitate synchronized predictions across varied scales.
Collectively, these benchmarks validate AMG's effectiveness in dynamic unstructured meshes, underlining its high precision and versatility in complex, real-world simulation scenarios. This robust performance underscores AMG's sophisticated capacity to navigate intricate interactions and continual changes.

\subsubsection{Three-Dimensional Benchmark Insights}
The Shape-Net Car benchmark involves predicting two attributes across a large mesh size of 32,186 points spread over three dimensions. The Deforming Plate benchmark presents a dynamic three-dimensional challenge, where the mesh evolves over time, adding another layer of complexity. 
In addressing these challenges, AMG leverages a combination of a graph specifically designed to handle arbitrary geometries, and multi-scale and physics graphs that effectively manage features across different scales. This integrated approach significantly enhances AMG's performance, leading to a substantial reduction in prediction errors by 35.44\%. This improvement not only highlights the model’s scalability but also its adeptness in processing volumetric data, effectively handling the intricacies of varied spatial and temporal scales, thereby markedly reducing prediction errors.

\subsubsection{Comparison with Graph-Based Methods}
Our analysis reveals that graph-based models such as MeshGraphNets (MGN) and Graph Neural Operator (GNO) generally underperform in static benchmarks like Shape-Net Car and Poisson, where the scenarios are dominated by low-frequency features. These models are better suited to dynamic benchmarks such as Deforming Plate, which involve varying shapes; however, their performance still falls short of expectations.
To overcome these limitations, AMG incorporates global and physics graphs. The global graph helps in capturing broader spatial relationships that are crucial for static environments, while the physics graph integrates fundamental physical laws and principles, enhancing the model's ability to accurately predict behaviors in dynamic scenarios. This strategic use of multiple graph types enables AMG to significantly outperform traditional graph-based methods across a range of benchmarks, effectively addressing both static and dynamic challenges.

\subsubsection{Comparison with Transformer-Based Methods}
Transformer-based models such as Transolver and GNOT previously set the standard for state-of-the-art performance in handling complex data structures. Despite their strengths, AMG has demonstrated superior performance by integrating specialized graph techniques that enhance feature processing capabilities across various scales. 
Utilizing the graph neural operator's ability to effectively manage local high-frequency features, combined with the global graph and physics graph's adeptness at capturing global low-frequency features, AMG has significantly outperformed these transformer-based models. This remarkable improvement showcases AMG's enhanced capability to address a broader range of dynamic and static phenomena more accurately than the existing state-of-the-art models.

\begin{table}[ht]\small
\centering
\caption{Results on Weather Forecasting.}
\label{table:weather}
\renewcommand\arraystretch{0.6}
    \resizebox{0.98\linewidth}{!}{
    \begin{tabular}{c|ccc}
       \toprule
       \multirow{2}{*}{Model} & \multicolumn{3}{c}{$L_{2}$ Error} \\ 
       \cmidrule(lr){2-4}
       & Temperature & Wind U & Wind V \\
       \midrule
       MLP & 0.0194 & 0.3198 & 0.3585 \\
       U-Net & 0.0113 & 0.2279 & 0.3108 \\
       \midrule
       MGN & 0.0186 & 0.3533 & 0.3559 \\
       GNO & 0.1173 & 0.3405 & 0.3896 \\
       \midrule
       ONO & 0.0156 & 0.2535 & 0.2914 \\
       GNOT & 0.0364 & 0.7954 & 1.0206\\
       LSM & 0.0164 & \underline{0.2095} & \underline{0.2435}\\
       Transolver & \underline{0.0077} & 0.2155 & 0.2762 \\
       \midrule
       \textbf{AMG (Ours)} & \textbf{0.0067} & \textbf{0.2020} & \textbf{0.2335} \\
       Promotion & $\downarrow$\textbf{13.08\%} & $\downarrow$\textbf{3.62\%} & $\downarrow$\textbf{4.11\%}\\
       \bottomrule
    \end{tabular}}
\end{table}

\subsubsection{Experiment on Real-World Dataset}
Our objective was to predict the next hour's temperature and wind components (U-component and V-component). The data was sampled from a reduced grid of the original, capturing 11,088 gridpoints, to balance computational demand with meaningful analysis. As illustrated in Table \ref{table:weather}, AMG outperforms other methods across all metrics, particularly excelling in dynamic conditions, such as predicting wind directions and speeds. This success underscores AMG's integration of multiscale graphs and dynamic graph attention mechanisms, enhancing its predictive accuracy in complex, variable scenarios.

These results affirm the robustness of AMG across diverse computational settings, from static and dynamic meshes to two-dimensional and three-dimensional spaces. The model's adeptness at navigating various mesh types and geometries, combined with its capability to manage intricate local and global interactions, establishes AMG as a powerful solution for solving PDEs in a wide range of applications. This comprehensive performance, particularly evident in real-world dynamic scenarios like weather forecasting, demonstrates AMG's potential to deliver precise, reliable predictions across different scales and conditions.

\begin{table}[ht]\small
\centering
\caption{Ablation Study on Graphs.}
\label{table:ablation}
\renewcommand\arraystretch{0.8}
\begin{sc}
    \renewcommand{\multirowsetup}{\centering}
    \resizebox{0.98\linewidth}{!}{
    \begin{tabular}{c|ccc}
       \toprule
       \multirow{2}{*}{Configuration} & \multicolumn{3}{c}{$L_{2}$ Error} \\ 
       \cmidrule(lr){2-4}
       & pressure & velocity x & velocity y \\
       \midrule
       w/o Local Graph & 0.0101 & 0.0076 & 0.0136 \\
       w/o Global Graph & 0.0135 & 0.0084 & 0.0149 \\
       w/o MultiScale Graph & 0.0087 & 0.0070 & 0.0137 \\
       w/o Physics Graph & 0.0598 & 0.0238 & 0.0622 \\
       \midrule
       \textbf{Baseline (AMG)} & \textbf{0.0050} & \textbf{0.0038} & \textbf{0.0078} \\
       \bottomrule
    \end{tabular}}
\end{sc}
\end{table}

\subsection{Model Analysis}
\subsubsection{Ablation Studies on Graph Configurations}
We conducted ablation studies to determine the contribution of different graph configurations—Local, Global, Multiscale, and Physics—to our model's performance. These studies, summarized in Table \ref{table:ablation}, show the impact of each graph type on the accuracy of predictions for pressure, velocity X, and velocity Y.

\begin{itemize}
    \item \textbf{Without Local Graph:} Removing this graph increased L2 errors, especially in velocity Y, underscoring its role in capturing local interactions.
    \item \textbf{Without Global Graph:} The absence led to higher L2 errors, highlighting its importance for contextual relationship understanding.
    \item \textbf{Without Multiscale Graph:} Its removal caused minor performance drops, indicating its role in linking local and global insights.
    \item \textbf{Without Physics Graph:} This had the most significant impact, emphasizing its necessity for incorporating physical principles.
\end{itemize}

The baseline AMG configuration outperformed all variations, illustrating the critical need for an integrated approach to handle complex simulations effectively.

\subsubsection{Hyperparameter Analysis}
Our comprehensive hyperparameter study, summarized in Table \ref{table:hp}, identifies optimal configurations that significantly enhance model performance across various metrics. The analysis revealed that simply increasing the number of layers, node numbers, or head numbers does not consistently reduce $L_{2}$ errors, indicating that optimal hyperparameter settings are crucial for maximizing performance. 
These results demonstrate that effective hyperparameter tuning is crucial for deploying AMG in diverse computational environments. By strategically selecting settings tailored to specific tasks, we can refine the model’s architecture, thereby enhancing both performance and operational efficiency. The insights gained from this study are instrumental in fine-tuning the model to achieve superior performance across different applications.

\begin{table}[ht]\small
\centering
\caption{Hyperparameter Study on Cylinder Flow.}
\label{table:hp}
\renewcommand\arraystretch{0.8}
\begin{sc}
    \renewcommand{\multirowsetup}{\centering}
    \resizebox{0.98\linewidth}{!}{
    \begin{tabular}{c|c|ccc}
       \toprule
       \multirow{2}{*}{Type} & \multirow{2}{*}{Configuration} & \multicolumn{3}{c}{$L_{2}$ Error} \\ 
       \cmidrule(lr){3-5}
       & & pressure & velocity x & velocity y \\
       \midrule
       \multirow{4}{*}{Layer Number} & layer=1 & 0.0148 & 0.0104 & 0.0188 \\
       & layer=2 & 0.0106 & 0.0073 & 0.0151 \\
       & \textbf{layer=3} & \textbf{0.0050} & \textbf{0.0038} & \textbf{0.0078}\\
       & layer=4 & 0.0096 & 0.0068 & 0.0141 \\
       \midrule
       \multirow{4}{*}{Local Node Number} & n=256 & 0.0096 & 0.0068 & 0.0141 \\
       & n=512 & 0.0107 & 0.0082 & 0.0148 \\
       & \textbf{n=1024} & \textbf{0.0050} & \textbf{0.0038} & \textbf{0.0078} \\
       & n=2048 & 0.0114 & 0.0077 & 0.0149 \\
       \midrule
       \multirow{3}{*}{Global Sample Ratio} & r=12.5\% & 0.0166 & 0.0109 & 0.0178 \\
       & \textbf{r=25\%} & \textbf{0.0050} & \textbf{0.0038} & \textbf{0.0078}\\
       & r=50\% & 0.0091 & 0.0069 & 0.0128 \\
       \midrule
       \multirow{4}{*}{Physical Node Number} & p=8 & 0.0171 & 0.0121 & 0.0230 \\
       & p=16 & 0.0142 & 0.0104 & 0.0204 \\
       & \textbf{p=32} & \textbf{0.0050} & \textbf{0.0038} & \textbf{0.0078}\\
       & p=64 & 0.0159 & 0.0116 & 0.0222 \\
       \midrule
       \multirow{4}{*}{Head Number} & h=1 & 0.0139 & 0.0092 & 0.0185 \\
       & h=4 & 0.0096 & 0.0066 & 0.0121 \\
       & \textbf{h=8} & \textbf{0.0050} & \textbf{0.0038} & \textbf{0.0078}\\
       & h=12 & 0.0096 & 0.0065 & 0.0122 \\
       \midrule
       \multirow{4}{*}{Global Graph Degree} & global k=2 & 0.0131 & 0.0095 & 0.0193 \\
       & \textbf{global k=4} & \textbf{0.0050} & \textbf{0.0038} & \textbf{0.0078} \\
       & global k=6 & 0.0090 & 0.0065 & 0.0115\\
       & global k=8 & 0.0179 & 0.0113 & 0.0233\\
       \midrule
       \multirow{4}{*}{Local Graph Degree} & local k=2 & 0.0152 & 0.0103 & 0.0176 \\
       & local k=4 & 0.0138 & 0.0111 & 0.0218 \\
       & \textbf{local k=6} & \textbf{0.0050} & \textbf{0.0038} & \textbf{0.0078}\\
       & local k=8 & 0.0101 & 0.0071 & 0.0150 \\
       \bottomrule
    \end{tabular}}
\end{sc}
\vspace{-10pt}
\end{table}

\subsubsection{Multi-Scale Sampling}
\begin{figure}[ht]
    \centering
    \subfigure[]{\includegraphics[width=0.48\linewidth]{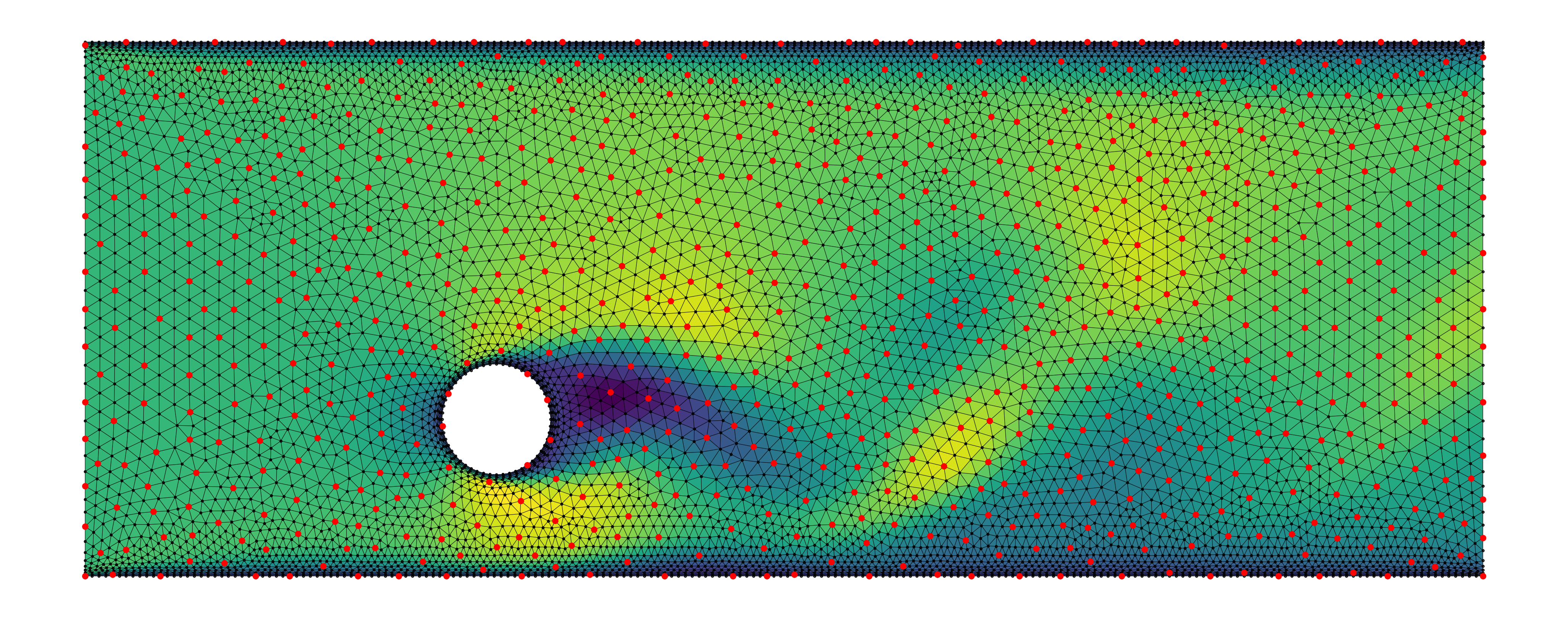}}\quad
    \subfigure[]{\includegraphics[width=0.48\linewidth]{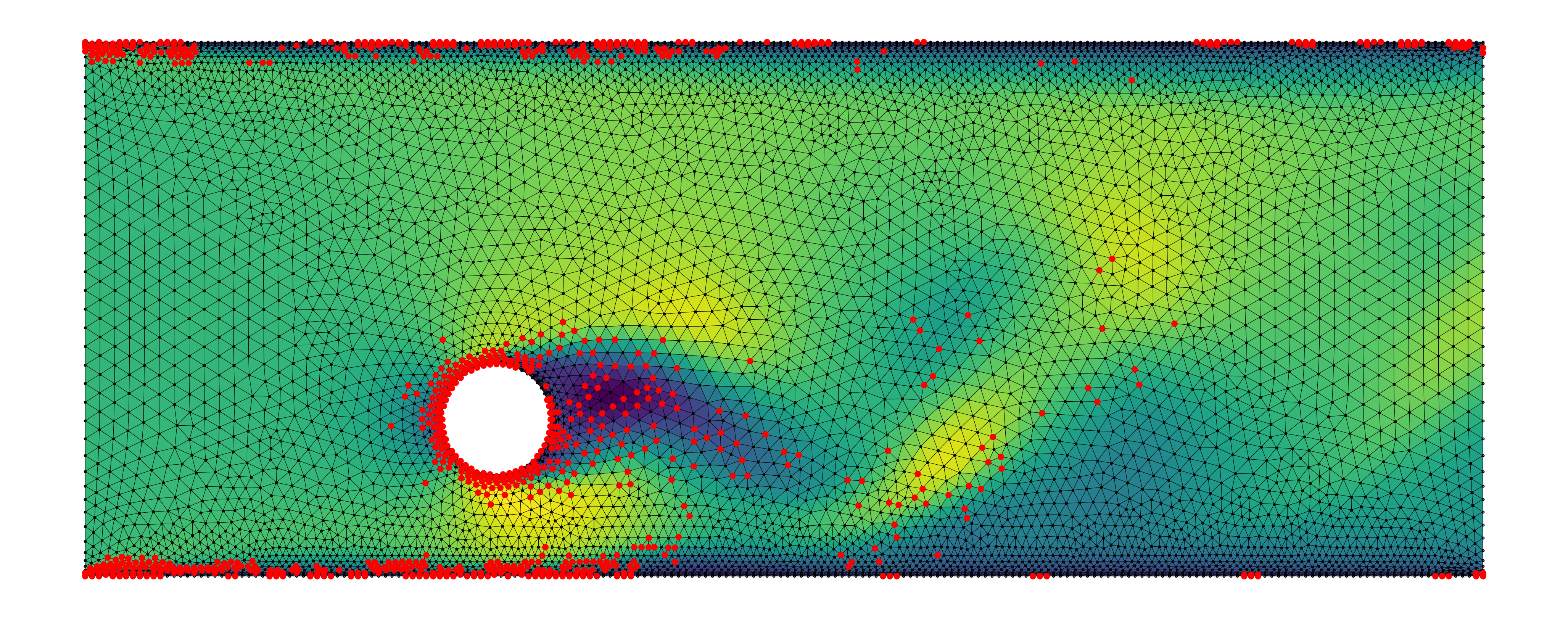}}\\
    \subfigure[]{\includegraphics[width=0.48\linewidth]{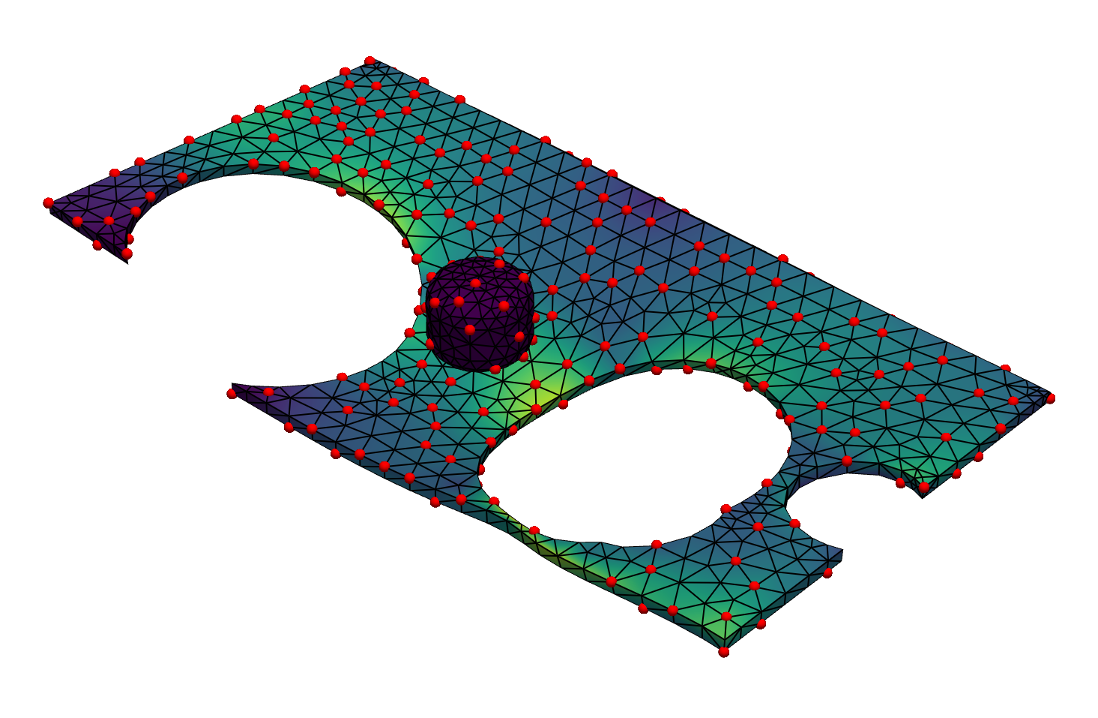}}\quad
    \subfigure[]{\includegraphics[width=0.48\linewidth]{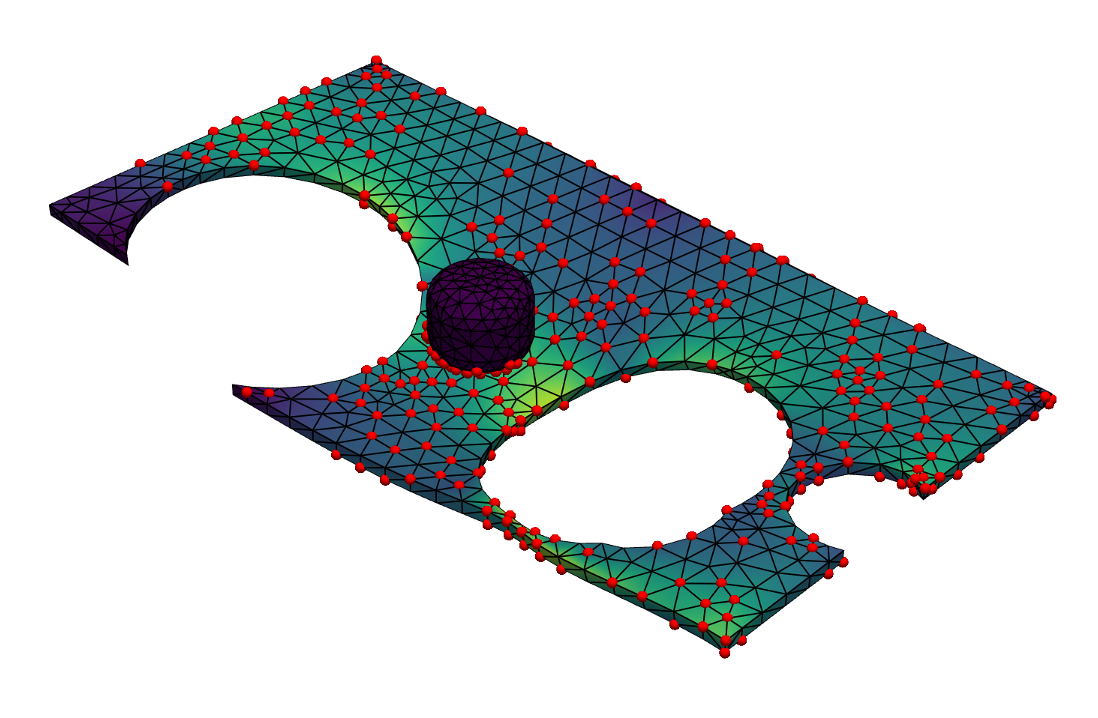}}
    \vspace{-10pt}
    \caption{Illustration of global sampling and local sampling in Cylinder Flow (a, b) and Deforming Plate (c, d).}
    \Description{}
    \label{fig:sample}
    \vspace{-10pt}
\end{figure}

Multi-Scale Sampling is a critical technique in our methodology, enabling the AMG model to effectively capture and integrate diverse spatial information. This approach involves two distinct sampling strategies: global sampling and local sampling. Global sampling (Figure \ref{fig:sample}(a,c) focuses on capturing broad-scale data integration across the entire domain, while local sampling (Figure \ref{fig:sample}(b,d)) targets high-resolution detail capture within specific regions of interest. These techniques collectively enhance the model's ability to process and analyze data from complex PDE systems with varying dynamics and geometric irregularities.

\section{Related Work}
\subsection{Neural Operators}
Neural Operators utilize deep neural networks to efficiently solve complex PDE systems by learning mappings between two functional spaces \cite{23:no}. Typically modeled as kernel integral operators, various parameterization strategies have been developed, including the widely recognized Fourier Neural Operators (FNO) which integrate Fast Fourier Transform (FFT) \cite{21:fno}.

\subsubsection{Graph-based Neural Operators}
Graph-based neural operators such as GNO \cite{20:GNO} utilize Message Passing Graph Networks to process data in irregular domains but face challenges like prolonged training and difficulty in capturing global low-frequency features \cite{22:NKN, 23:WNO_in_computational_mechanics, 23:GINO, 20:MGNO, 23:LSM}. MGNO \cite{20:MGNO} attempts to overcome these by using multi-scale graphs, yet its reliance on pre-defined matrix decomposition restricts flexibility with dynamic meshes and complex geometries. Recent models like GINO and GeoFNO \cite{23:GINO, 24:GeoFNO} combine graph-based strategies with Fourier Neural Operators to adapt irregular data into uniform grids, aiming to resolve some of the enduring challenges of GNO. However, these advancements still contend with foundational issues inherent in the original graph-based methods.

\subsubsection{Transformer-based Neural Operators}
Transformer-based neural operators utilize attention mechanisms to encode operators in latent spaces effectively. The Galerkin Transformer \cite{21:Choose} uses a Galerkin-type linear attention for operator parameterization, while ONO \cite{23:ONO} introduces an orthogonal attention module inspired by Mercer's theorem to refine the kernel integral. OFormer \cite{23:OFormer} employs a standard transformer structure with cross-attention, encoding both input functions and query locations into a latent ODE representation. GNOT \cite{23:GNOT} and the state-of-the-art Transolver \cite{24:Transolver} further enhance these concepts. GNOT employs heterogeneous normalized cross-attention for uniform feature handling, and Transolver leverages physics-aware tokens for more precise operator learning. Despite their advantages, transformer models can struggle with detailed point-level attention, which may result in missing high-frequency information in complex PDE environments.

\subsubsection{Multi-level Neural Operators}
Multi-level neural operators are adept at representing PDE behavior across varying scales, enhancing performance significantly. The Multiwavelet Transform (MWT) \cite{21:MWT} uses hierarchical multiwavelet projections to handle kernel operator singularities and signal fluctuations. Advanced multigrid techniques like MgNO \cite{24:MgNO} and M2NO \cite{24:M2NO} incorporate the V-cycle of traditional multigrid methods into neural operator learning, with M2NO also integrating features from algebraic multigrid and multiresolution wavelet transform for superior performance.
While these methods excel in managing resolution scales, they face challenges in irregular domains and can be computationally demanding. This underscores the need for further refinement to ensure both efficiency and broad applicability in complex PDE scenarios.

\subsection{Geometric Deep Learning}
Geometric Deep Learning is pivotal for managing irregular geometries, leveraging advances outlined in \cite{16:GeometricDL}. Graph Neural Networks (GNNs) are central to this domain, utilizing kernels on connected graphs for effective representation learning, as demonstrated in \cite{17:InductiveRL,19:GraphU,21:learningMS}. Additionally, PointNet \cite{16:PointNet,17:PointNetDH} and Point Transformer \cite{20:PointT} effectively handle scattered point clouds.

For PDE learning, the MAgNet architecture \cite{22:MAgNet} merges coordinate-based methods with GNNs, enhancing performance in irregular meshes. Implicit neural representations for PDE dynamics forecasting, notably \cite{23:Continuous} and \cite{23:CROM}, enhance the modeling capabilities in dynamic systems. The work of \cite{23:NeuralImplicitFlow} provides a mesh-agnostic approach to dimensionality reduction, particularly beneficial for handling spatio-temporal data in continuous domains. The Graph Neural Operator (GINO) \cite{23:GINO} and Geo-FNO \cite{24:GeoFNO} adapt irregular grids into more uniform latent structures, optimizing the application of Fourier neural operators. However, transformer-based models like LSM \cite{23:LSM}, GNOT \cite{23:GNOT}, and Transolver \cite{24:Transolver} still face significant challenges with arbitrary geometries and high-frequency detail capture, which are crucial for accurately solving PDEs across various scales.

\section{Conclusion and Future Work}
This paper introduces AMG, a novel Multi-Graph neural operator method designed to efficiently solve Partial Differential Equations (PDEs) on Arbitrary geometries. AMG harnesses advanced graph-based techniques combined with dynamic graph attention mechanisms within an innovative GraphFormer architecture. This approach enables the precise handling of diverse spatial domains and intricate data interdependencies, setting it apart from previous methods constrained to uniform grids.

Our extensive evaluations across various benchmarks confirm AMG's superior performance and scalability compared to existing state-of-the-art solutions. These results demonstrate AMG's potential to significantly advance the computational mathematics field, especially in applications involving complex and non-linear systems.

Looking ahead, we aim to explore the possibilities of large-scale pre-training of AMG, similar to the development of foundation models in other domains of AI. Such advancements could establish a new paradigm for PDE solving, potentially leading to more generalized and robust methods capable of tackling a broader range of scientific and engineering challenges. This direction not only promises to enhance the model's performance but also explores its adaptability and applicability to increasingly complex scenarios.

Here’s a more concise and consistent version of the acknowledgements:

\begin{acks}
Wei Wang was supported by the Guangdong Provincial Key Lab of Integrated Communication, Sensing and Computation for Ubiquitous Internet of Things (No. 2023B1212010007), and the Guangzhou Municipal Science and Technology Projects (Nos. 2023A03J0003, 2023A03J0013, and 2024A03J0621). Zhilu Lai was supported by the Guangzhou-HKUST(GZ) Joint Funding Grant (No. 2023A03J0105), the Guangdong Provincial Key Lab of Integrated Communication, Sensing and Computation for Ubiquitous Internet of Things (No. 2023B1212010007), and the Guangzhou Municipal Science and Technology Project (No. 2023A03J0011).
\end{acks}

\bibliographystyle{ACM-Reference-Format}
\balance
\bibliography{reference}

\appendix

\section{Sampling Algorithm Details} \label{appendix:sample}
\subsection{Farthest Point Sampling Algorithm} \label{appendix:fps}
Farthest Point Sampling (FPS) \cite{20:PointT} is a technique used to select a subset of points from a larger set in such a way that the selected points are as far apart from each other as possible. This method is particularly useful in scenarios where uniform coverage across the entire dataset is crucial. As detailed in Algorithm~\ref{al:fps}, given a set of input points $\{\mathbf{x}_1, \mathbf{x}_2, \ldots, \mathbf{x}_n\}$, FPS iteratively selects points $\{\mathbf{x}_{i_1}, \mathbf{x}_{i_2}, \ldots, \mathbf{x}_{i_m}\}$, ensuring that each new point $\mathbf{x}_{i_j}$ is the farthest away from previously selected points $\{\mathbf{x}_{i_1}, \mathbf{x}_{i_2}, \ldots, \mathbf{x}_{i_{j-1}}\}$. This strategy enhances coverage and is particularly advantageous over random sampling, especially in applications where data distribution is non-uniform.

\RestyleAlgo{ruled}
\begin{algorithm}
\caption{Farthest Point Sampling Algorithm}
\label{al:fps}
\SetKwInOut{Input}{Input}
\SetKwInOut{Output}{Output}

\Input{Coordinates Set $C$ of shape $(N, D)$, where $N$ is the number of nodes and $D$ is the dimensionality; $n$, the number of nodes to sample.}
\Output{$sampledIndices$}

$minDistances \gets \{\infty\}$\;
$sampledIndices \gets \{-1\}$\;

$current \gets \text{randint}(0, N-1)$ 
$sampledIndices[0] \gets current$\;

\For{$i \gets 1$ \KwTo $n - 1$}{
    \For{$j \gets 0$ \KwTo $N-1$}{
        $distance \gets \text{computeDistance}(C[j], C[current])$\;
        \If{$distance < minDistances[j]$}{
            $minDistances[j] \gets distance$\;
        }
    }
    $current \gets \text{argmax}(minDistances)$\;
    $sampledIndices[i] \gets current$\;
}

\Return{$sampledIndices$}\;
\end{algorithm}

Unlike convolutional neural networks (CNNs), which process data without considering underlying spatial distributions, FPS takes a data-dependent approach to create receptive fields, thus preserving important spatial hierarchies within the data. This method is highly effective for constructing graphs or other data structures where the geometric arrangement of the points is significant.

\subsection{Local Sampling Algorithm} \label{appendix:local}

Local sampling is essential for constructing graphs that accurately capture the immediate neighborhood dynamics of nodes, which is vital for handling complex local interactions in PDE-solving environments. This method utilizes a high-frequency indicator (see Eq.\eqref{eq:HF}) to selectively sample detailed nodes, ensuring that each local graph includes only the most informative nodes. The sampling process begins by generating a feature map that highlights regions with high detail. Using the \texttt{topK} function, the top $n$ nodes with the highest feature values within each batch are selected. Subsequently, these selected nodes are connected using a K-nearest neighbors (\texttt{knn}) approach based on their spatial positions, effectively capturing the local structural information.

Algorithm \ref{al:local} outlines the Local Sampling Algorithm, which operates as follows: given node features $X$, coordinates $C$, batch indices, the number of nodes to sample $n$, and the number of nearest neighbors $k$, the algorithm processes each batch by computing the feature map, selecting the top $n$ nodes using the \texttt{topK} function, and constructing edges between these nodes using the \texttt{knn} method. This targeted sampling and graph construction enable the model to adapt to complex data variations and local interactions, thereby enhancing its ability to solve PDEs with high precision.

\begin{algorithm}
\caption{Local Sampling Algorithm}
\label{al:local}
\SetKwInOut{Input}{Input}
\SetKwInOut{Output}{Output}

\Input{
    $X$: Node Set of shape $(N, D_{x})$, where $N$ is the number of nodes and $D_{x}$ is the dimensionality of node features.\;
    $C$: Coordinates Set of shape $(N, D_{pos})$, where $D_{pos}$ is the dimensionality of positions.\;
    $batch$: Array of batch indices corresponding to each node.\;
    $n$: Number of nodes to sample per batch.\;
    $k$: Number of nearest neighbors to consider.
}
\Output{$sampledIndices$, $sampledEdgeIndex$}

$sampledIndices \gets \emptyset$\;
$sampledEdgeIndex \gets \emptyset$\;

$featureMap \gets \text{computeFeatureMap}(X, C, batch)$\;

\For{$i \gets 1$ \KwTo $|\text{unique}(batch)|$}{
    $mask \gets (batch = i)$\;
    
    $sampledIndices[i] \gets \text{topK}(featureMap[mask], n)$\;
    
    $sampledEdgeIndex[i] \gets \text{knn}(C[mask], C[mask], k)$\;
}

\Return{$sampledIndices$, $sampledEdgeIndex$}\;
\end{algorithm}

\section{Proof of Theorem \ref{th:integral}} \label{appendix:proof}
The theorem is established by demonstrating that the graph attention mechanism employed in the GraphFormer can be formalized as a Monte-Carlo approximation of an integral operator \cite{21:Choose, 23:no, 24:Transolver}.

\begin{proof}
Given an input function $a: \Omega \to \mathbb{R}^d$ on a domain $\Omega \subset \mathbb{R}^d$, the integral operator $\mathcal{G}$ can be formalized as:
\begin{equation} \label{eq:integral}
    \mathcal{G}a(\mathbf{x}) = \int_{\Omega} \kappa(\mathbf{x}, \boldsymbol{\xi}) a(\boldsymbol{\xi}) \, d\boldsymbol{\xi},
\end{equation}
where $\mathbf{x}\in\Omega\subset\mathbb{R}^{d}$ and $\kappa(\cdot,\cdot)$ denotes the kernel function defined on $\Omega$. We define the kernel for the graph attention mechanism as follows:
\begin{equation} \label{eq:kernel}
    \begin{aligned}
    \kappa(\mathbf{x},\boldsymbol{\xi})=&\left(\int_{\Omega}\mathrm{exp}\left(\mathbf{a}^\top \left[ \mathbf{W}a(\boldsymbol{\xi}) \| \mathbf{W}a(\boldsymbol{\xi'}) \right]\right)d\boldsymbol{\xi'}\right)^{-1}\\
    &\mathrm{exp}\left( \mathbf{a}^\top \left[ \mathbf{W}a(\mathbf{x}) \| \mathbf{W}a(\boldsymbol{\xi}) \right] \right)\mathbf{W},
    \end{aligned}
\end{equation}
where \( \mathbf{a} \in \mathbb{R}^{2d} \), \( \mathbf{W} \in \mathbb{R}^{d \times d} \) and \(\| \) denotes vector concatenation. 
Using the softmax normalizing function across the domain $\Omega$, assuming $\Omega$ carries a uniform measure for simplicity in the Monte-Carlo approximation.

Approximating the integrals by a finite sum over the neighborhood $\mathcal{N}(\mathbf{x})$ of $\mathbf{x}$ in a graph:
\begin{equation}
\begin{aligned}
    &\int_{\Omega}\mathrm{exp}\left(\mathbf{a}^\top \left[ \mathbf{W}a(\boldsymbol{\xi}) \| \mathbf{W}a(\boldsymbol{\xi'}) \right]\right)d\boldsymbol{\xi'}\\
    \approx& \frac{|\Omega|}{|\mathcal{N}({\mathbf{x}})|}\sum_{i\in\mathcal{N}(\mathbf{x})}\mathrm{exp}\left( \mathbf{a}^\top \left[ \mathbf{W}a(\mathbf{x}_{i}) \| \mathbf{W}a(\boldsymbol{\xi})\right] \right),
\end{aligned}
\end{equation}
we then use this to express $\mathcal{G}a(\mathbf{x})$ of Eq.\eqref{eq:integral} as a weighted sum:
\begin{equation}
    \mathcal{G}a(\mathbf{x}) \approx \frac{\sum_{i \in \mathcal{N}(\mathbf{x})} \exp(\mathbf{a}^\top \left[ \mathbf{W}a(\mathbf{x}) \| \mathbf{W}a(\mathbf{x}_i) \right]) \mathbf{W}a(\mathbf{x}_i)}{\sum_{j \in \mathcal{N}(\mathbf{x})} \exp(\mathbf{a}^\top \left[ \mathbf{W}a(\mathbf{x}) \| \mathbf{W}a(\mathbf{x}_j) \right])},
\end{equation}
which is the computation for the graph attention mechanism of GraphFormer defined in Eq.\eqref{eq:alpha}-\eqref{eq:ga}, showing its equivalence to a learnable integral operator on the domain $\Omega$.
\end{proof}

\section{Details for Benchmarks} \label{appendix:bm}
We evaluate AMG across various datasets, including those specifically generated for this study using COMSOL \cite{98:COMSOL}.

\begin{figure}[ht]
    \centering
    \subfigure[Unstructured Mesh]{\includegraphics[width=0.3\linewidth]{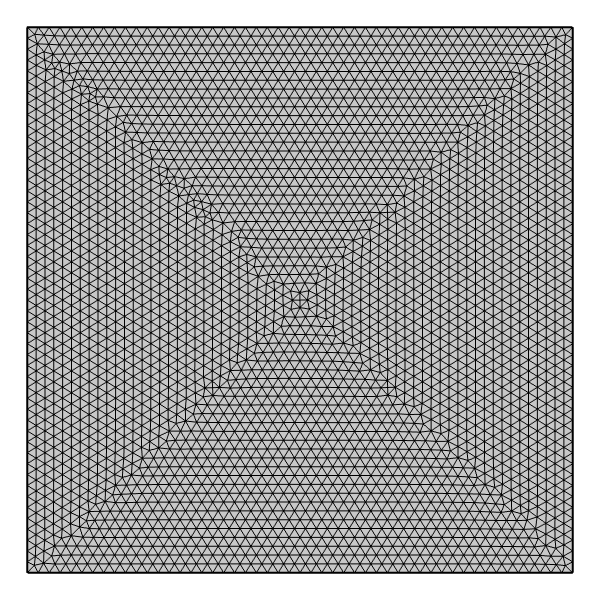}}
    \subfigure[Input: $f$]{\includegraphics[width=0.34\linewidth]{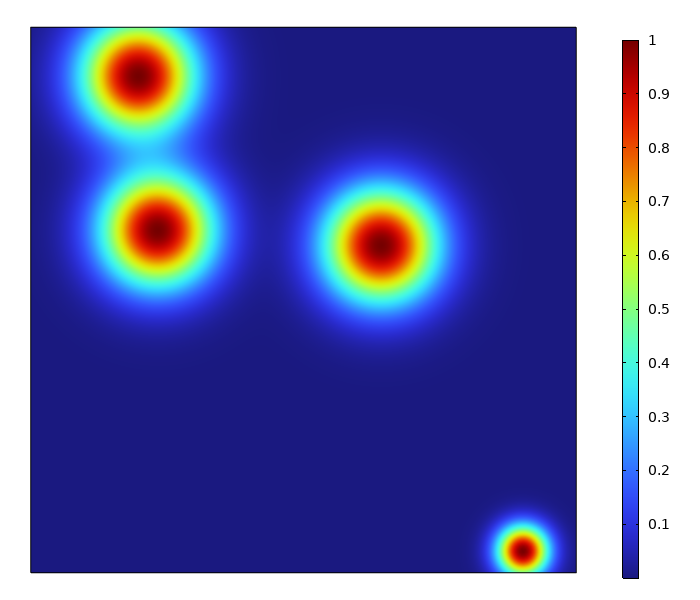}}
    \subfigure[Output: $u$]{\includegraphics[width=0.34\linewidth]{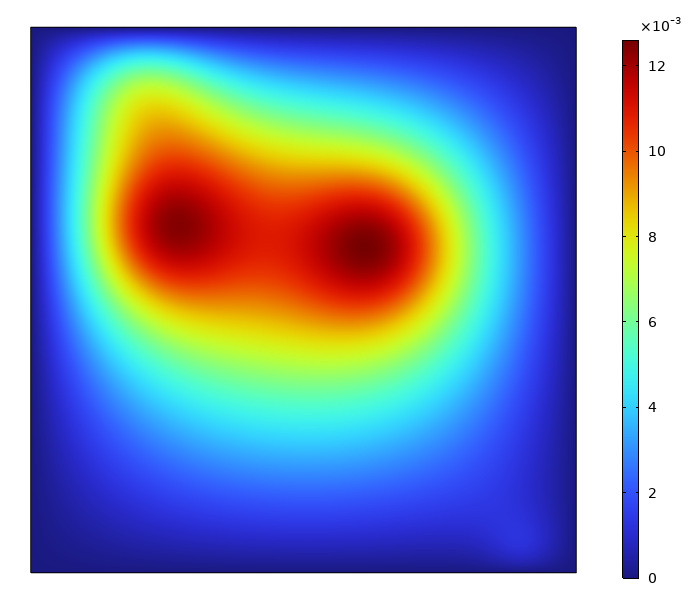}}
    \vspace{-10pt}
    \caption{Visualization of a random sample from the Poisson dataset.}
    \Description{An example from the Poisson dataset illustrating both the input function and the output solution.}
    \label{fig:poisson}
    \vspace{-10pt}
\end{figure}

\subsection{Poisson Equation}
The Poisson equation with Dirichlet boundary conditions is studied:
\begin{align}
    -\Delta u &= f, \quad \text{in } \Omega = [0,1]^2, \\
    u &= 0, \quad \text{on } \partial \Omega,
\end{align}
where \(f\) consists of a Gaussian superposition, with parameters \(\mu_{x,i}, \mu_{y,i} \sim \text{U}(0,1)\) and \(\sigma_i \sim \text{U}(0.025, 0.1)\). The dataset includes 4000 training, 500 validation, and 500 test samples.

\begin{figure}[ht]
    \centering
    \subfigure[Input: Past Fluid velocity]{\includegraphics[width=0.45\linewidth]{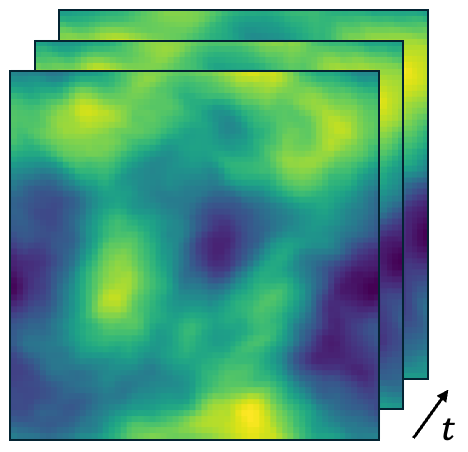}}
    \hspace{0.5cm} 
    \subfigure[Output: Future fluid velocity]{\includegraphics[width=0.45\linewidth]{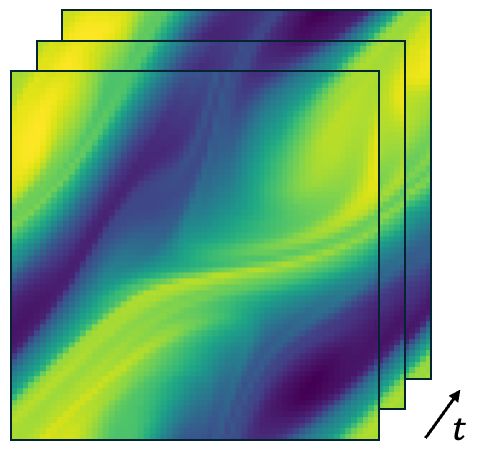}}
    \vspace{-10pt}
    \caption{Visualization of the Navier-Stokes dataset for a sample simulation.}
    \Description{}
    \label{fig:ns}
    \vspace{-10pt}
\end{figure}

\subsection{Navier-Stokes Equation}
This section explores the Navier-Stokes equations in vorticity form on a unit torus:
\begin{align}
    \partial_t \omega + \mathbf{u} \cdot \nabla \omega - \nu \Delta \omega &= f, \\
    \nabla \cdot \mathbf{u} &= 0,
\end{align}
The dataset employs a \(64 \times 64\) grid, with periodic boundary conditions. It comprises 9600 training, 1200 validation, and 1200 test samples.

\begin{figure}[ht]
    \centering
    \subfigure[Mesh of the sample]{\includegraphics[width=0.48\linewidth]{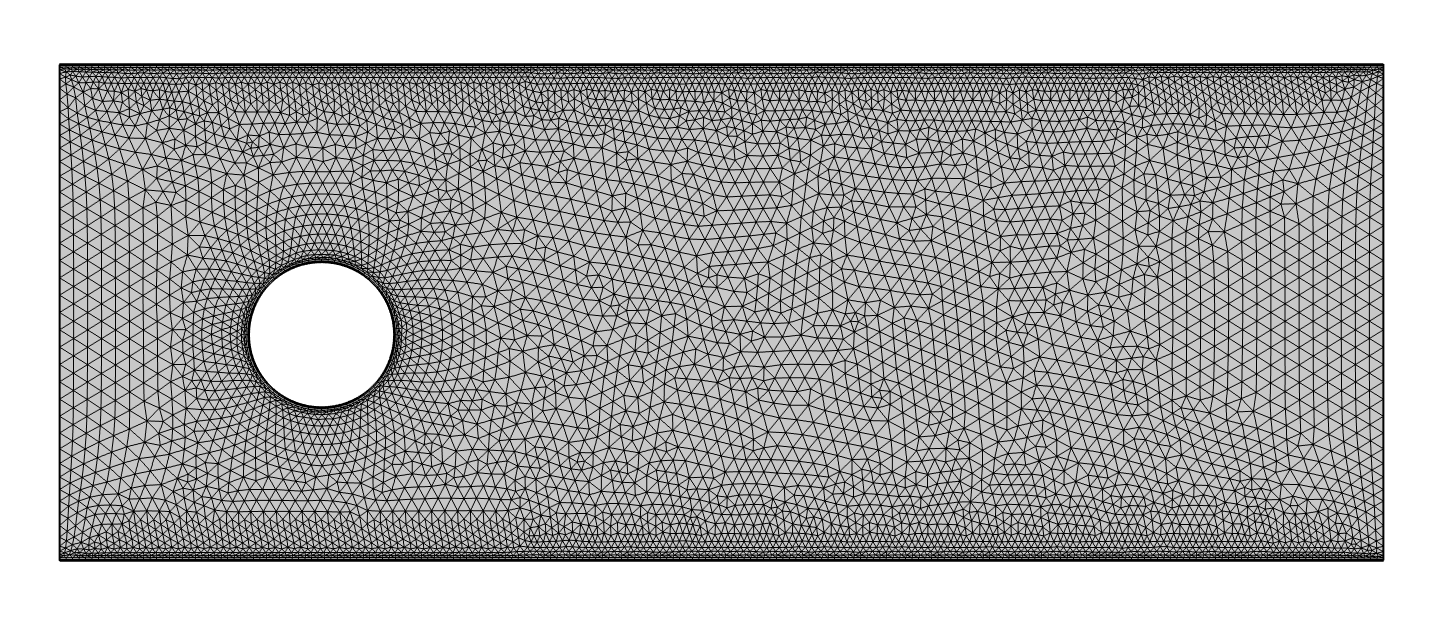}}
    \subfigure[Velocity field in future]{\includegraphics[width=0.45\linewidth]{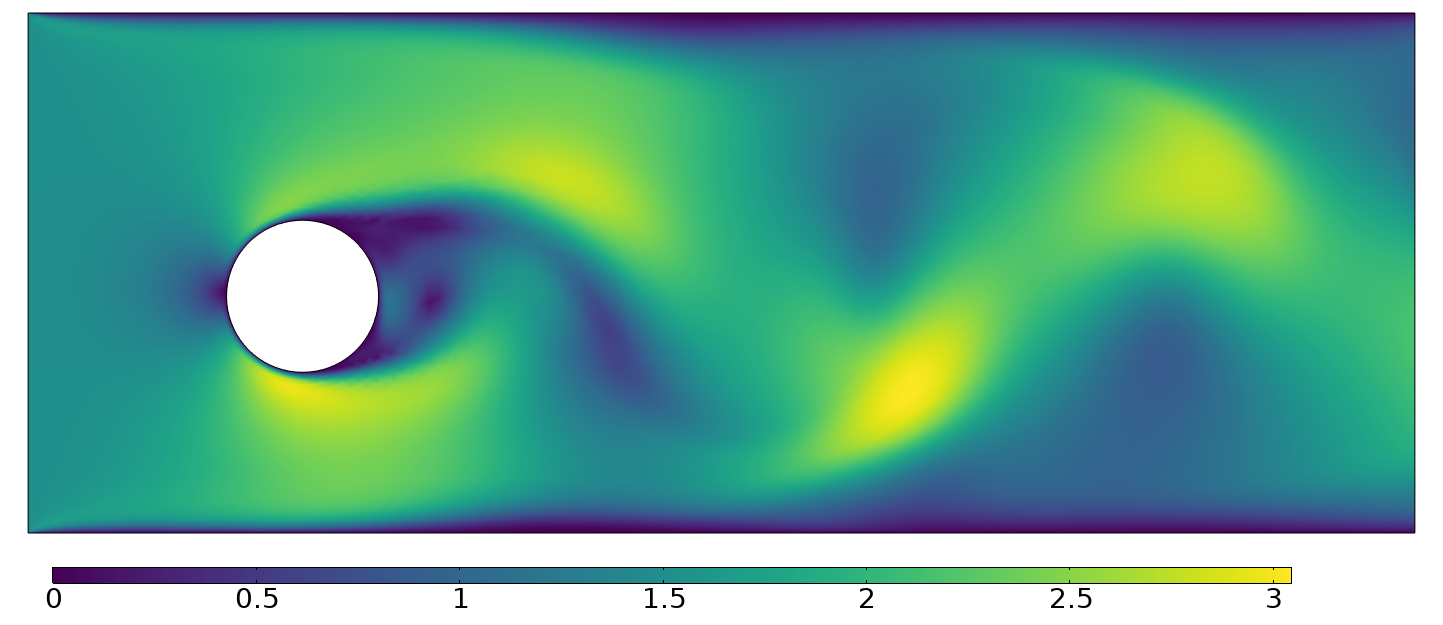}} 
    \vspace{-10pt}
    \caption{Visualization of a sample from the Cylinder Flow.} 
    \label{fig:cylinder} 
    \Description{An example from the Cylinder Flow dataset illustrating the mesh setup and velocity field dynamics.}
    \vspace{-10pt}
\end{figure}

\subsection{Cylinder Flow}
This dataset captures incompressible fluid dynamics around a 2D circular cylinder within a channel:
\begin{align}
    \nabla \cdot \mathbf{u} &= 0, \\
    \partial_t \mathbf{u} + (\mathbf{u} \cdot \nabla) \mathbf{u} &= \nu \nabla^2 \mathbf{u} - \frac{1}{\rho} \nabla p,
\end{align}
with boundary conditions set for velocity and pressure. It features 100 snapshots per case, with 7600 training, 1000 validation, and 1000 test samples.

\begin{figure}[ht]
    \centering
    \subfigure[Input: Fluid quantity at the current timestep]{\includegraphics[width=0.45\linewidth]{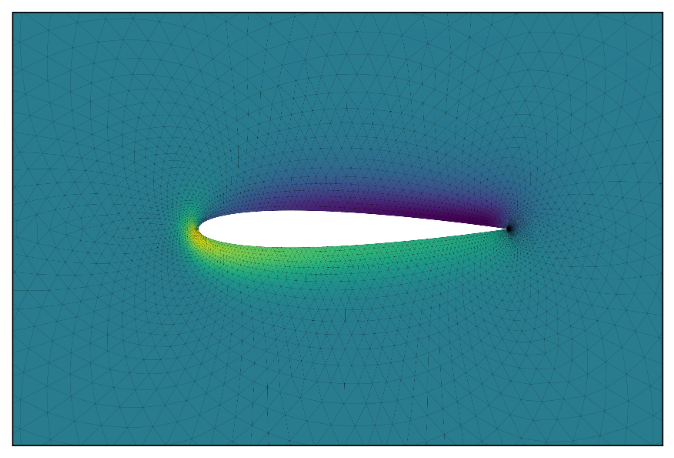}}
    \hspace{0.5cm}
    \subfigure[Output: Predicted fluid quantity at the next timestep]{\includegraphics[width=0.45\linewidth]{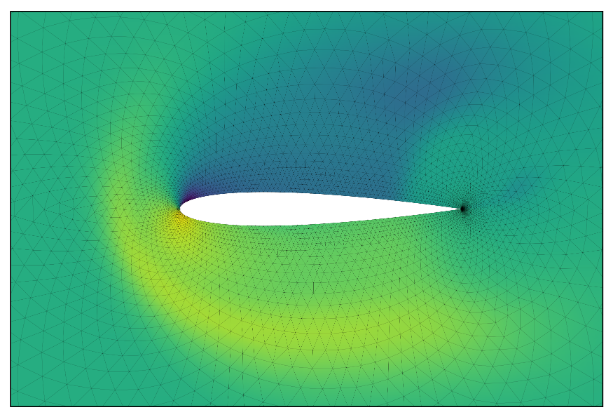}}
    \vspace{-10pt}
    \caption{Visualization of the Airfoil dataset illustrating a typical simulation snapshot.}
    \Description{}
    \label{fig:airfoil}
    \vspace{-10pt}
\end{figure}

\subsection{Airfoil}
Simulating subsonic airflow over an airfoil with the 2D compressible Euler equations:
\begin{align}
        \partial_t{\rho} + \nabla \cdot (\rho \mathbf{u}) &= 0,\\
        \partial_t (\rho \mathbf{u}) + \nabla \cdot (\rho \mathbf{u} \otimes \mathbf{u} + p \mathbf{I}) &= 0,
\end{align}
The simulation uses an unstructured mesh over 10 timesteps. This dataset consists of 10,000 training, 1,000 validation and testing samples each.

\begin{figure}[ht]
    \centering
    \subfigure[Input: Initial Stress]{\includegraphics[width=0.45\linewidth]{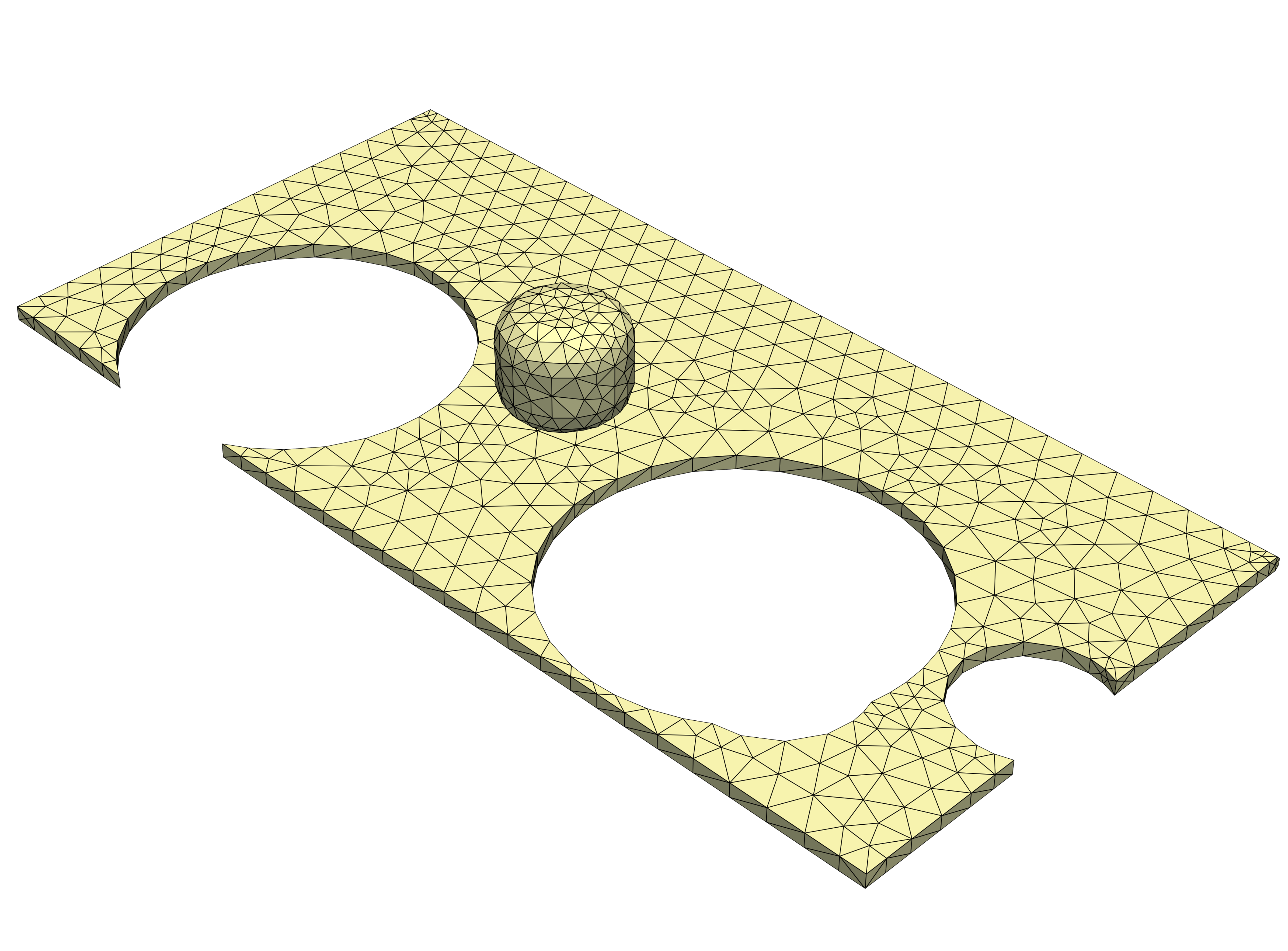}}
    \hspace{0.5cm} 
    \subfigure[Output: Predicted Stress]{\includegraphics[width=0.45\linewidth]{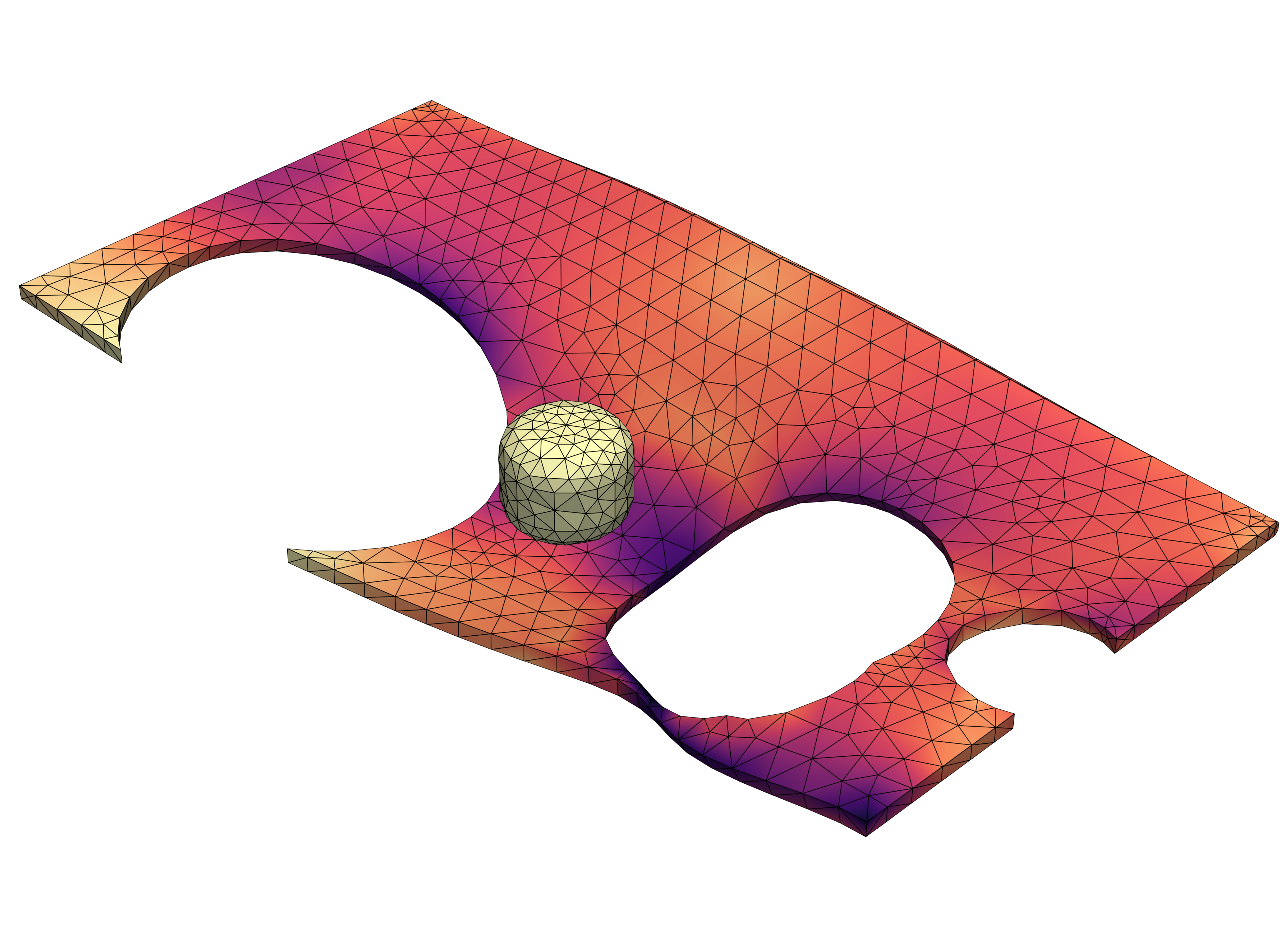}}
    \vspace{-10pt}
    \caption{Visualization of the input and output of Von Mises stress of a deforming plate.}
    \label{fig:deforming}
    \Description{}
    \vspace{-10pt}
\end{figure}

\subsection{Deforming Plate}
A 3D dynamic simulation of a deformable plate is modeled in hyperelastic material properties:
Data spans 50 timesteps, comprising 24,000 training samples, with 2,000 for validation and testing.

\begin{figure}[ht]
    \centering
    \subfigure[Input: Fluid quantity at the current timestep]{\includegraphics[width=0.45\linewidth]{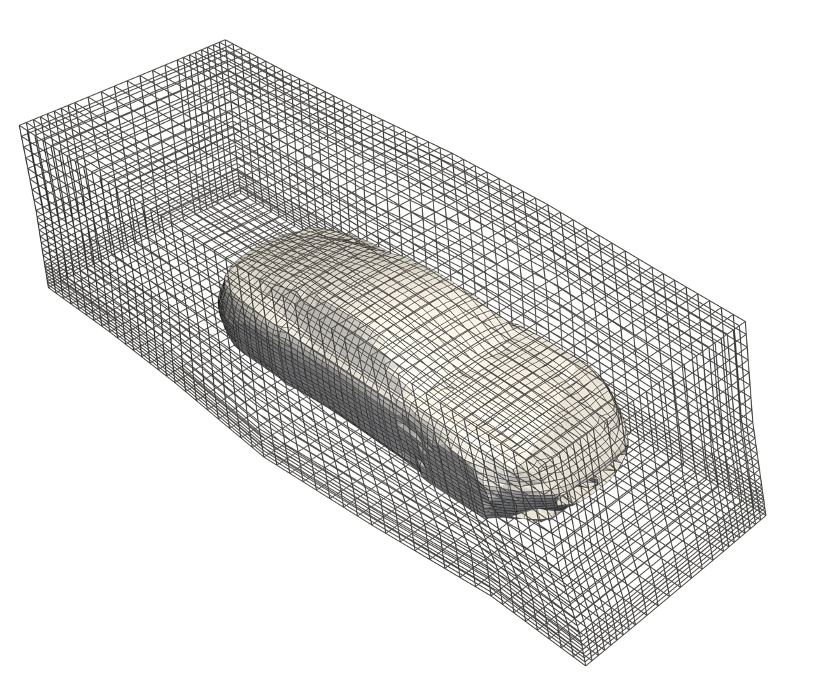}}
    \hspace{0.5cm}
    \subfigure[Output: Predicted fluid quantity at the next timestep]{\includegraphics[width=0.45\linewidth]{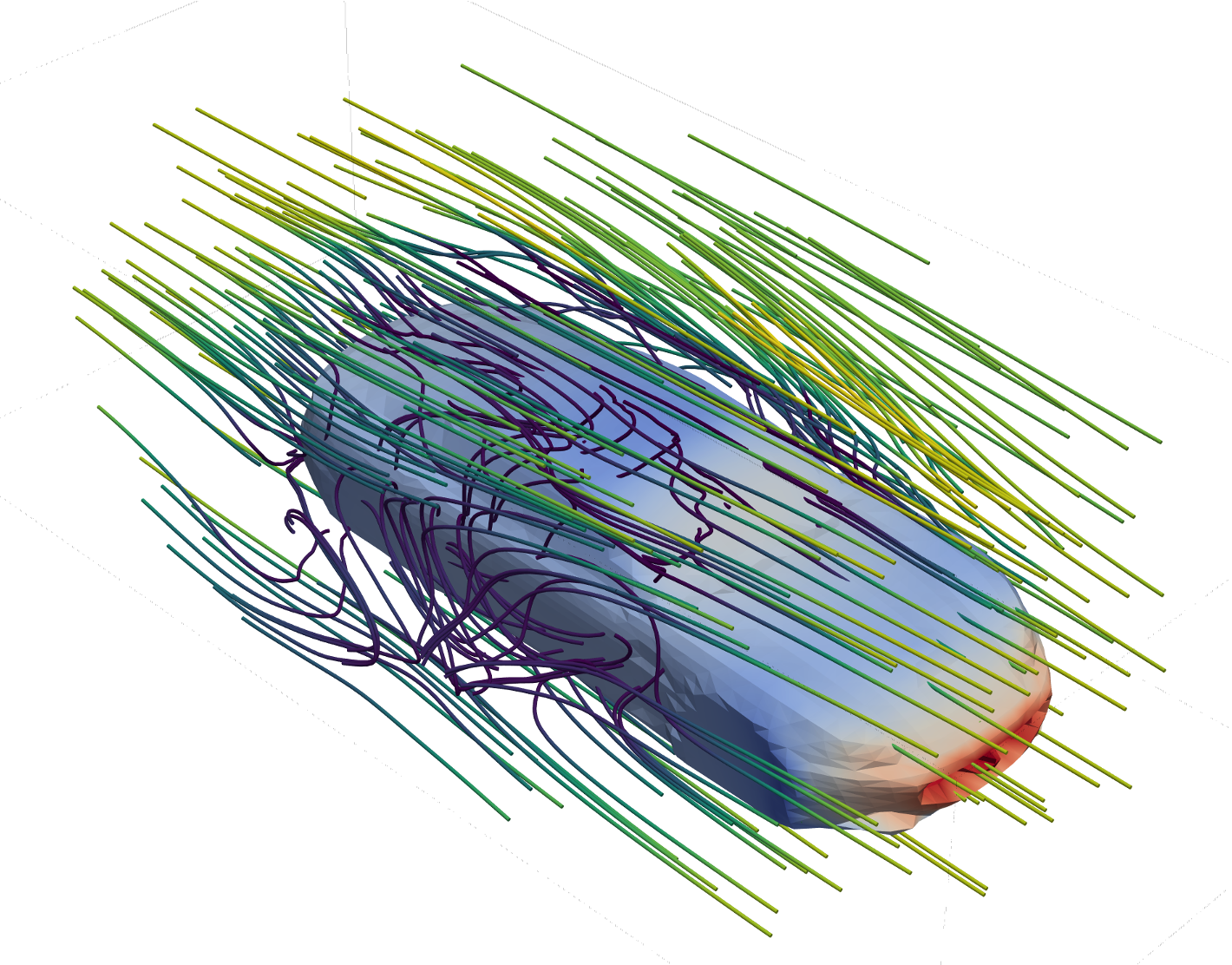}}
    \vspace{-10pt}
    \caption{Visualization of the Shape-Net Car dataset.}
    \Description{}
    \label{fig:car}
    \vspace{-10pt}
\end{figure}

\subsection{Shape-Net Car}
Drag coefficient estimation for cars simulated at 72 km/h, with 889 total samples: 690 for training, 99 for validation, and 100 for testing. Predicts velocity and pressure to compute drag coefficients.

\end{document}